\documentclass{article} % For LaTeX2e
\usepackage{gy_template,times}

\usepackage{amsmath,amsfonts,bm}

\def\eqref#1{equation~\ref{#1}}
\def\1{\bm{1}}

\DeclareMathAlphabet{\mathsfit}{\encodingdefault}{\sfdefault}{m}{sl}
\SetMathAlphabet{\mathsfit}{bold}{\encodingdefault}{\sfdefault}{bx}{n}

\PassOptionsToPackage{numbers, compress}{natbib}
\definecolor{goodblue}{rgb}{0.21,0.49,0.74}
\usepackage[pagebackref,breaklinks,colorlinks,allcolors=goodblue]{hyperref}

\usepackage{etoolbox}
\usepackage{graphicx}
\usepackage{caption}
\usepackage{subcaption}
\usepackage{wrapfig}
\usepackage{multicol}
\usepackage{multirow}
\usepackage{booktabs} 
\usepackage{overpic}
\usepackage[table]{xcolor}
\definecolor{mycommentcolor}{RGB}{0,0,255} 

\definecolor{light_yellow}{RGB}{255,243,194}
\definecolor{orange}{RGB}{255,200,100}
\definecolor{light_gray}{RGB}{237,237,237}
\definecolor{mid_gray}{RGB}{205,205,205}    

\newcommand{\best}{\cellcolor{mid_gray}}
\newcommand{\second}{\cellcolor{light_gray}}
\newcommand{\bestbox}{\colorbox{mid_gray}{best}}
\newcommand{\secondbox}{\colorbox{light_gray}{second}}

\newcommand{\name}{\textsc{FoundAD}}

\usepackage{pifont}
\newcommand{\cmark}{\ding{51}} % ✓
\newcommand{\xmark}{\ding{55}} % ✗
\usepackage{xcolor}

\usepackage{placeins}

\title{Foundation Visual Encoders Are Secretly Few-Shot Anomaly Detectors}

\author{%
  Guangyao Zhai$^{1,2}$ \thanks{The first two authors contribute equally.} \hspace{2ex}%
  Yue Zhou$^{1,2}$ \footnotemark[1] \hspace{2ex}%
  Xinyan Deng$^{1}$\hspace{2ex}%
  Lars Heckler$^{1,3}$\\[0.5ex] %
  {Nassir Navab}$^{1,2}$\hspace{2ex}%
  {Benjamin Busam}$^{1,2}$\\[2ex]%
  $^1$Technical University of Munich\hspace{1.2ex}%
  $^2$Munich Center for Machine Learning\hspace{1.2ex}%
  $^3$MVTec Software GmbH\\[0.5ex]
  \texttt{\{firstname.lastname,b.busam\}@tum.de}\hspace{1.5ex}
  \texttt{lars.heckler@mvtec.com}
}

\begin{document}
\maketitle

\begin{abstract}
Few-shot anomaly detection streamlines and simplifies industrial safety inspection. However, limited samples make accurate differentiation between normal and abnormal features challenging, and even more so under category-agnostic conditions. Large-scale pre-training of foundation visual encoders has advanced many fields, as the enormous quantity of data helps to learn the general distribution of normal images.
We observe that the anomaly amount in an image directly correlates with the difference in the learnt embeddings and utilize this to design a few-shot anomaly detector termed \name{}.
This is done by learning a nonlinear projection operator onto the natural image manifold.
The simple operator acts as an effective tool for anomaly detection to characterize and identify out-of-distribution regions in an image.
Extensive experiments show that our approach supports multi-class detection and achieves competitive performance while using substantially fewer parameters than prior methods.
Backed up by evaluations with multiple foundation encoders, including fresh DINOv3, we believe this idea broadens the perspective on foundation features and advances the field of few-shot anomaly detection. Our code is at { \small \url{ https://github.com/ymxlzgy/FoundAD}}.
\end{abstract} 
\section{Introduction}
\label{sec:intro}
\begin{wrapfigure}{r}{0.5\textwidth}
  \vspace{-3em}
  \captionsetup{type=figure}
  \includegraphics[width=\linewidth]{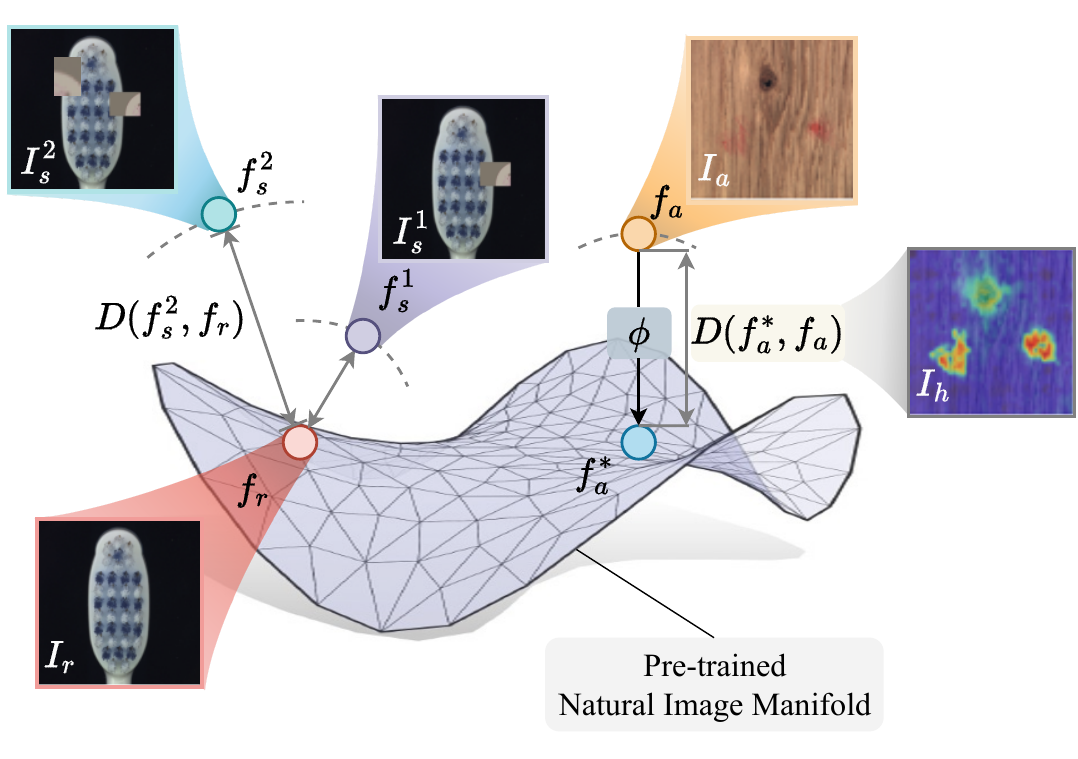}
  \vspace{-1.8em}
  \captionof{figure}{\small \textbf{Manifold Projection}. Large training sets enable foundation models to learn the manifold of natural images (illustrated schematically as a 2D surface), which lies in a higher-dimensional feature space. Normal images such as $I_r$ are embedded onto this manifold. Images with anomalies ($I_s^1$, $I_s^2$) lie further away from this manifold. The distance $D\left( f_s^i, f_r \right)$ correlates with the pixel amount of the anomaly in the image. We learn a non-linear projection operator $\phi$ that projects the embedding $f_a$ of an anomalous image $I_a$ onto its corresponding normal feature $f_a^\ast$. Feature comparison enables few-shot anomaly detection $I_h$.}
  \label{fig:teaser}
  \vspace{-1.5em}
\end{wrapfigure}
The variability of images is large.
Understanding general concepts from pixel data is therefore by design a very complex problem.
Foundation models~\citep{zagoruyko2016wide,caron2021emerging,zhai2023sigmoid} have provided a significant leap forward to image understanding in generalized contexts across many tasks.
The design of training paradigms and the advances in computational resources enabled the computer vision community to learn powerful visual encoders whose feature embeddings are adjustable to many downstream tasks~\citep{Dinov2}.
These foundation visual encoders are trained to encode samples from the core distribution of normal natural images as illustrated by the natural image manifold in~\autoref{fig:teaser}.
\begin{figure*}[t]
    \centering
    \includegraphics[width=\textwidth]{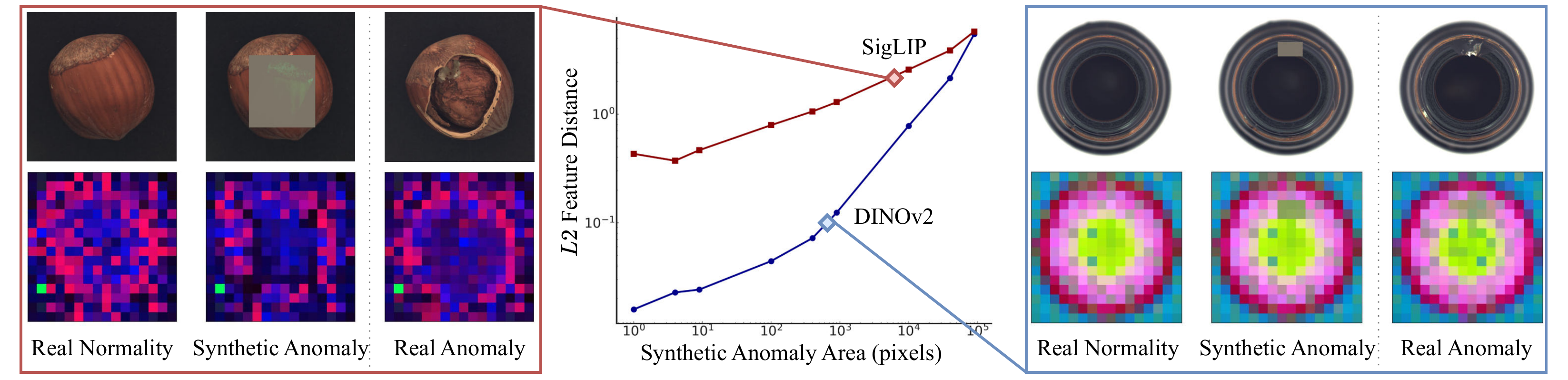}
    \caption{ \small \textbf{Correlation of Anomaly Area with Feature Distance}. Two foundation encoders under different paradigms are shown. \emph{Upper left/right}: Real images with corresponding synthetic and real anomalies. \emph{Lower left/right}: Coloured PCA visualizations of their embedded features using SigLIP~\citep{zhai2023sigmoid} (left) and DINOv2~\citep{Dinov2} (right). \emph{Center}: L2-feature distance of embeddings for synthetic anomalies of increasing pixel amount on a real image. A clear correlation is visible for both foundation models.}
    \label{fig:motivation}
    \vspace{-0.5mm}
\end{figure*}
For some vision tasks, specifically the tail end of this distribution away from the heavy core is important~\citep{bergmann2019mvtec,visa}.
Industrial inspection, for instance, demands robust anomaly detection techniques that can operate effectively even when limited annotated data is available, as collecting large datasets in production is not only costly, but also impractical. This applies especially to defective samples, where the types of defects that might potentially occur are unknown prior to the launch of production.
For these setups, unsupervised few-shot anomaly detection is a highly attractive approach, since it only requires a small amount of defect-free samples for training. However, the scarcity of anomalous samples generally hampers the possibility to learn comprehensive  representations~\citep{chen2020simple}, such that traditional methods face the challenge of distinguishing subtle differences between normal and abnormal features.

Studying the embedding structure of foundation models in light of this task reveals an interesting property:
The amount of anomalous area in an image directly corresponds to the feature distance in the embedding space as shown in~\autoref{fig:motivation}.
This shows that foundation encoders distinguish anomaly from normality, {i.e.}, they \emph{“secretly detect”} the defective regions.
We attribute this to the learning signal from natural images that stipulates the creation of a natural image manifold in the embedding space of a foundation model. Moving away from this structure correlates to shifting towards the tail end or outside the general distribution of normal images.

We leverage this observation in our work to design \name{}, a few-shot anomaly detector that utilizes the image embeddings of a foundation visual encoder.
% Inspired by joint embedding predictive architectures (JEPA)~\ci'te'p{jepa,ijepa}, which learn robust, consistent representations by enforcing cross-view prediction consistency, we design a nonlinear feature projector that effectively transitions features toward the normal manifold (cf.~\autoref{fig:teaser}).
We utilize foundational models to encode both anomalous and normal features, where structural anomalies are primitively synthesized by CutPaste \citep{CutPaste}.
The encoder is frozen to bootstrap the learning process by reusing the strong semantic and geometric understanding.
Then, we employ a nonlinear projector to learn the necessary feature mapping with only a minimal few-shot demonstrations, such as a single sample.
We validate \name{} on multiple foundation visual encoders, including DINO series~\citep {caron2021emerging,Dinov2,simeoni2025dinov3}, vision branches of vision-language models (VLMs)~\citep{zhai2023sigmoid,radford2021learning}, and a pre-trained convolutional neural network (CNN) \citep{zagoruyko2016wide}.
The experiments show that \name{} with DINOv3 achieves the best performance, and even it is trained on multiple classes with compact parameters, it is superior to methods that particularly focus on each class~\cite{li2024promptad}, and it performs competitively among large-scale few-shot anomaly detectors~\cite{lv2025oneforall,zhang2025towards}.

Conceptually, \name{} is inspired by predictive embedding approaches such as JEPA~\citep{jepa,ijepa} and SimSiam~\citep{chen2021exploring}, which capture representation dependencies between paired inputs by operating purely in latent space, {eschewing} the need for pixel-level observation reconstruction. In contrast to these methods actively training encoders, we keep the foundation encoders frozen to enjoy the natural image manifold, and instead solely train a simple network to adapt and project the pretrained embeddings exclusively and effectively for the few-shot anomaly detection (cf.~\autoref{fig:teaser}).
The entire pipeline is fast, lightweight, and easy to train. More importantly, our findings suggest that rethinking the use of foundational visual encoders can extend their applicability to complex anomaly detection tasks \emph{without} additional design complexity, such as conventional textual prompts for assistance~\citep{li2024promptad,lv2025oneforall,zhang2025towards}.

In summary, the contributions of our work are threefold:
\textbf{(i)} We reveal the {correlation} of {embedding distance and} {anomaly amount} in images for foundation visual encoders.
\textbf{(ii)} Based on point (i), we introduce a {feature projection method} to efficiently {discriminate between anomaly and normality} in the embedding space. The nonlinear projector is lightweight and can be trained with minimal demonstrations.
\textbf{(iii)} Comprehensive experiments demonstrate that the {performance} of our pipeline {surpasses state-of-the-art multi-class methods}. More importantly, it reveals that foundation visual features \emph{without} textual assistance suffice for few-shot anomaly detection.

\begin{figure*}[t]
    \centering
    \includegraphics[width=\linewidth]{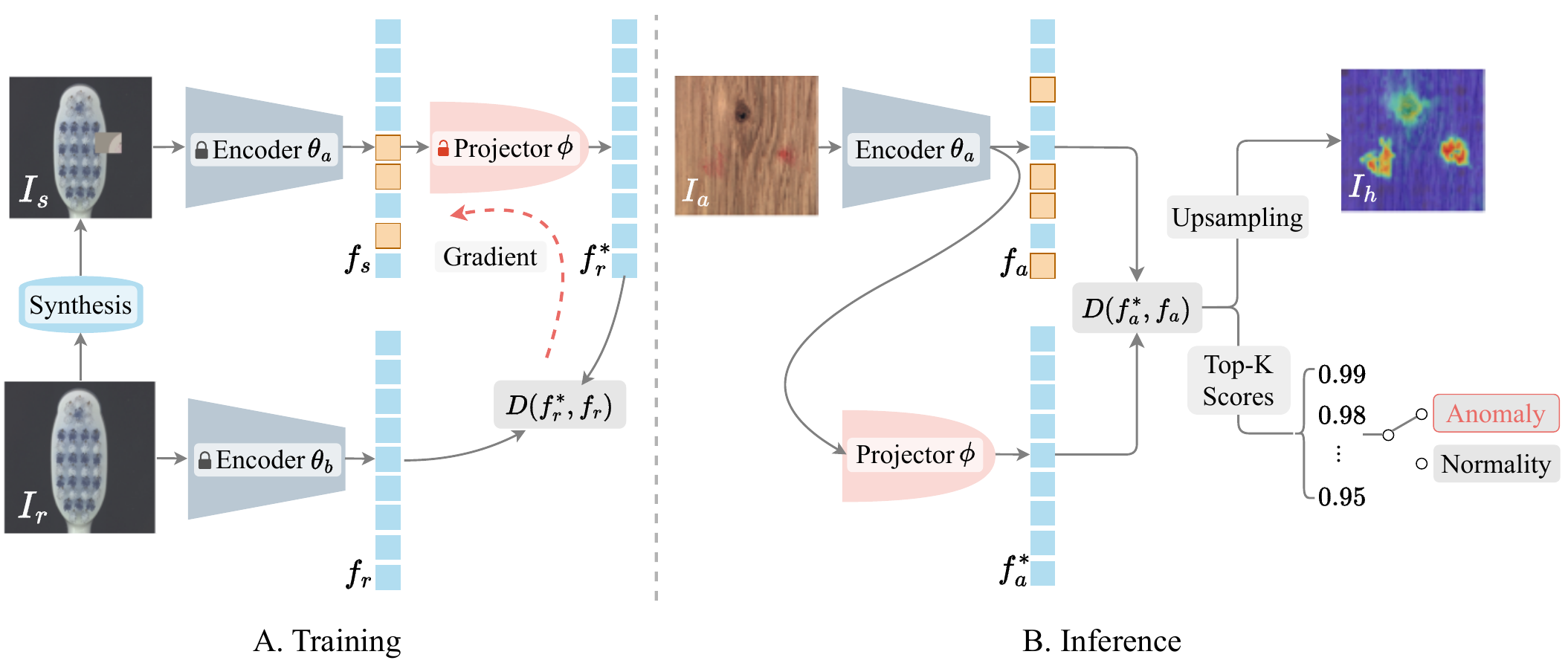} 
    \caption{\small \textbf{A. Training pipeline.} Normal training images $I_r$ are first processed by the anomaly synthesis module to generate augmented samples $I_s$. Feature embeddings of the augmented image and the original image are extracted by the \emph{Anomaly-Aware Encoder} \( \theta_a \) and the \emph{Reference Encoder} \( \theta_b \), respectively ($\theta_a=\theta_b=\theta$). The \emph{Manifold Projector} $\phi$ is trained to map the feature embeddings $f_s$ of the synthesized anomalous image towards the normal feature $f_r$. The training objective is to minimize the distance $D\left( f_r^\ast, f_r \right)$ between the projected feature $f_r^\ast$ and the reference feature $f_r$. \textbf{B. Inference pipeline.} During inference, an input image is processed by AE to extract feature embeddings $f_a$, which are then projected by the Projector to $f_a^\ast$. The anomaly score $D\left( f_a^\ast, f_a \right)$ for each patch is computed. We aggregate the Top-K highest patch-level anomaly scores and generate an anomaly heatmap $I_h$ by upsampling to the original image resolution.}
    \label{fig:pipeline}
    \vspace{-3mm}
\end{figure*}
\section{Methodology}
\label{sec:AD-JEPA}
The overall architecture is illustrated in \autoref{fig:pipeline}. The network is designed to project the latent representation of either an abnormal image or a normal image to the natural image manifold learned by a foundation model. The projected representation is compared to the original representation, where large differences indicate the occurrence of anomalies. The key components of the framework include an anomaly synthesis module, two identical encoders, and a projector.

\subsection{Anomaly Synthesis}
\label{syn}
To train the framework in an unsupervised setting, we utilize a structural anomaly synthesis module inspired by CutPaste \citep{CutPaste}. As shown in \autoref{fig:motivation}, synthesized anomalies exhibit noticeable differences from real anomalies at the pixel level, but these differences become less pronounced in the latent space. This suggests that a simple synthesis strategy is sufficient to drive anomalous features away from the natural image manifold, aligning with our projector’s training objective. To enhance the reality of synthetic anomaly, we constrain anomalies to foreground regions using adaptive threshold-based binarization \citep{zhang2024realnet, yang2023memseg}.

\subsection{Visual Manifold Projection}  
Our framework starts with two foundation visual encoders to extract latent representations from input images, as shown in \autoref{fig:pipeline}: the \emph{Anomaly-Aware Encoder} (AE), which processes synthesized abnormal images \( I_s \), and the \emph{Reference Encoder} (RE), which processes original normal images \( I_r \). Both encoders share the same parameters \( \theta \), ensuring consistency in feature extraction and aligning embeddings within the same latent space. This design keeps normal patches in synthesized images close to their counterparts in normal images, while highlighting discrepancies in anomalous regions.  

On top of these representations, we introduce the \emph{Manifold Projector}, a nonlinear module applied after AE that maps the anomaly-aware features toward the natural image manifold with only few-shot supervision. Since the features are tokenized, we implement the projector using a self-attention Vision Transformer (ViT). Each transformer block employs residual connections,  
\(
x_{\text{out}} = \operatorname{Attn}(x_{\text{in}}) + x_{\text{in}},
\)
to stabilize training and preserve input information. Given an anomaly image \( I_s \), the encoder yields an embedding \( f_s = \theta(I_s) \), which deviates from the normal embedding \( f_r \). The projector \( \phi \) then produces a corrected embedding \( f_r^* \) that aligns with the normal feature \( f_r \) on the natural image manifold. Unlike reconstruction-based approaches such as~\citep{yan2024anomalysd}, our method operates entirely in latent space, substantially reducing computational complexity.

% Unlike few-shot embedding-based methods requiring support images, it functions independently with only the query image

\subsection{Training}
The training process follows the pipeline illustrated in ~\autoref{fig:pipeline}.
During training, given an original normal image \( I_r \) in each iteration, we independently gate anomaly synthesis with a threshold \(\sigma\).
Let \(z \sim \mathrm{Bernoulli}(1-\sigma)\), and then
$I_s = (1-z)\, I_r + z\, \operatorname{Syn}(I_r),$ where \( \operatorname{Syn} \) denotes the synthesis operation described Sec.~\ref{syn}. 
We obtain feature embeddings using \( \theta \):
$f_s = \theta(I_s), f_r = \theta(I_r).$
Then the projector \( \phi \)  maps the synthesized anomaly feature \( f_s \) to the estimated normal feature:
$f_r^* = \phi(f_s).$
The training objective is to minimize the discrepancy between the projected feature and the reference normal feature via:
\begin{equation}
\label{loss}
    \mathcal{L} = {D}(f_r^*, f_r) = \frac{1}{N} \sum_{i=1}^{N} (f_{r,i}^* - f_{r,i})^2,
\end{equation}
where $\mathcal{L}$ is the $L2$ loss, and \(N\) is the number of embedded patches.

\subsection{Inference} 
Given a test image \( I_a \), its feature embedding is extracted using AE as \(f_a = \theta(I_a)\). The projector \( \phi \) is then applied to map the extracted feature embedding to the estimated feature on the natural image manifold $f_a^* = \phi(f_a)$.
We compute the patch-level anomaly score as the squared $L2$ distance between the extracted feature and its projected counterpart, turning \autoref{loss} to:
\begin{equation}
    S_{\text{patch}} = {D}(f_a^*, f_a) = \frac{1}{N} \sum_{i=1}^{N} (f_{a,i}^*-f_{a,i})^2.
\end{equation}
For image-level anomaly detection, we compute the average of the Top-K patch anomaly scores:
\begin{equation}
    S_{\text{image}} = \frac{1}{K} \sum_{i=1}^{K} S_{\text{patch}, i}.
\end{equation}
Finally, to generate pixel-level anomaly maps, the patch-level scores are upsampled to the original image resolution, highlighting regions with the most significant deviations.

\section{Experiments}
\label{sec:exp}
\subsection{Experimental Setup}
\paragraph{Datasets}
Following the protocol in IIPAD~\citep{lv2025oneforall}, we evaluate \name{} with various foundation encoders on two popular industrial anomaly detection datasets: MVTec-AD \citep{bergmann2019mvtec} and VisA \citep{visa}. MVTec-AD consists of $5,354$ high-resolution images across five texture and ten object categories, with $1,725$ for testing, covering various real-world defects such as contamination and structural deformations. VisA includes twelve object categories with $10,821$ images, featuring $9,621$ normal samples and $1,200$ anomalous images. It presents additional challenges due to complex structures, multi-instance objects, and diverse anomaly patterns, requiring models to generalize across varying levels of complexity.

\paragraph{Evaluation Metrics}
To comprehensively assess anomaly detection and localization performance, we employ three standard metrics: (1) Area Under the Receiver Operating Characteristic Curve (AUROC), which evaluates the model’s ability to distinguish normal and anomalous samples at both image and pixel levels; (2) Area Under the Precision-Recall Curve (AUPR), which is particularly effective for imbalanced datasets by emphasizing precision-recall trade-offs; and (3)Per-Region-Overlap (PRO), which measures region-level anomaly localization by computing the overlap between predicted and ground-truth anomalous regions. Curve areas involving the false positive rate are calculated only up to an FPR of $0.3$ \citep{bergmann2019mvtec}.
These metrics collectively ensure a robust evaluation of both detection accuracy and localization effectiveness.

\paragraph{Baseline}
We evaluate the method against state-of-the-art anomaly detection methods across multi/one-class settings. For few-shot and multi-class scenarios, we compare with FastRecon \citep{fang2023fastrecon}, AnomalySD \citep{yan2024anomalysd}, and IIPAD \citep{lv2025oneforall}.
%, and UniVAD \citep{gu2025univad}. 
Additionally, we include one-class methods such as WinCLIP \citep{jeong2023winclip}, InCTRL~\citep{zhu2024generalistanomalydetectionincontext}, AnomalyCLIP \citep{zhouanomalyclip}, PromptAD \citep{li2024promptad}, and LogSAD \citep{zhang2025towards}, as well as classical models like SPADE \citep{spade} and PatchCore \citep{PatchCore}, referring to results reported in IIPAD. 
% For full-shot multi-class benchmarks, we follow IIPAD \citep{lv2025oneforall} to compare against UniAD \citep{you2022unified}, OmniAL \citep{zhao2023omnial}, DiAD \citep{DiAD}, and HVQ-Trans \citep{lu2023hierarchical} in the Supplementary Material.

\paragraph{Implementation Details}
\name{} is compatible with various encoders, and we report the one backed by a pretrained DINOv3 ViT-B \citep{simeoni2025dinov3} here. For details about more encoders, please refer to the Supplementary Material~\ref{encoder_ablation_detail}. The input images from MVTec-AD and VisA are resized to \( 512 \times 512 \times 3 \) for consistency across experiments. The projector is implemented as a ViT with a depth of $6$ layers. For stable and efficient optimization, we utilize the Adam optimizer with a weight decay of \( 1 \times 10^{-4} \). The learning rate is set to \( 0.001\), empirically chosen to ensure convergence. 
% The number of training steps varies based on the available shots: 2,800 steps for 1-shot, 8,200 steps for 2-shot, 16,700 steps for 4-shot, and xxx steps for 8-shot experiments. 
All experiments are conducted on a single RTX 3090 GPU with a batch size of $8$. During training, we set the synthesis threshold \(\sigma\) as $0.5$. For inference, the image-level anomaly score is computed based on the Top-K highest patch anomaly scores, where \( K \) is set to $10$ for MVTec-AD and $6$ for VisA.

\subsection{Experimental Results}

For each few-shot setting, we conduct three different sample combinations, each drawn under a random seed. The averaged results on the MVTec-AD and VisA datasets across different few-shot settings are reported in \autoref{table:results_multi_class}. Some one-class-one-model full-shot methods (e.g., SPADE, PatchCore) from IIPAD are adapted to the multi-class-one-model few-shot setting for fair comparison. To further demonstrate the competitiveness of \name{}, \autoref{table:results_single_class} presents results under the one-class-one-model paradigm. Detailed numbers of each run are provided in Supplementary Material~\ref{multiple_run}. We also conduct ablation studies with different foundation visual encoders as backbones in \autoref{ablation_encoder}, with different DINOv3 layers in \autoref{ablation_layer}, and with different projector designs in \autoref{tab:Different_projector}. Finally, we show some typical failure cases in \autoref{fig:failure_case} to illustrate the limitations of \name{}.

\begin{table}[h]
    \centering
    \footnotesize
    \caption{ \small
    Comparison of results on MVTec-AD and VisA against various \textbf{multi-class-one-model} few-shot methods. Metrics include image-level AUROC (\%), AUPR (\%), and pixel-level AUROC (\%), PRO (\%). We color the \bestbox{} and the \secondbox{} in the k-shot setting.
    }
    \label{table:results_multi_class}
    \resizebox{0.95\textwidth}{!}{
    \begin{tabular}{llccccccccc}
    \toprule
    \multirow{2}{*}{\textbf{Shot}} & \multirow{2}{*}{\textbf{Method}} & \multirow{2}{*}{\textbf{w/o  Texts}}
      & \multicolumn{4}{c}{\textbf{MVTec-AD}} 
      & \multicolumn{4}{c}{\textbf{VisA}} \\
    \cmidrule(l{1em}r{1em}){4-7}
    \cmidrule(l{1em}r{1em}){8-11}
     &  & & {I-AUROC} & {AUPR} & {P-AUROC} & {PRO}
        & {I-AUROC} & {AUPR} & {P-AUROC} & {PRO} \\
    \midrule

    %--------- 1-shot + Multi-class ---------
    \multirow{9}{*}{\shortstack{1}}
    & SPADE        & \color{green!60!black}\cmark & 58.8 & 63.7 & 60.4 & 53.1 & 61.3 & 68.2 & 69.0 & 57.2 \\
    & PatchCore    & \color{green!60!black}\cmark & 63.7 & 81.2 & 83.9 & 72.7 & 58.9 & 62.8 & 76.7 & 64.3 \\
    & FastRecon    & \color{green!60!black}\cmark & 51.2 & 72.6 & 62.1 & 60.3 & 55.0 & 72.8 & 70.7 & 58.2 \\
    & WinCLIP      & \color{red!70!black}\xmark & 92.8 & 96.5 & 92.4 & 83.5 & 83.1 & 85.1 & 94.6 & 80.9 \\
    & PromptAD     & \color{red!70!black}\xmark & 86.3 & 93.4 & 91.8 & 83.6 & 80.8 & 83.2 & 96.3 & 82.2 \\
    & AnomalySD    & \color{red!70!black}\xmark & 93.6 & {96.9} & 94.8 & 89.2 & \second{86.1} & \second{89.1} & 96.5 & \second{93.9} \\
    & IIPAD        & \color{red!70!black}\xmark & \second{94.2} & \second{97.2} & \second{96.4} & \second{89.8} & 85.4 & 87.5 & \second{96.9} & 87.3 \\
    & \name{} \textbf{(Ours)} 
                   & \color{green!60!black}\cmark & \best96.1 & \best97.9 & \best96.8 & \best92.8 & \best92.6 & \best92.0 & \best99.7 & \best98.0\\
    \midrule

    %--------- 2-shot + Multi-class ---------
    \multirow{9}{*}{\shortstack{2}}
    & SPADE        & \color{green!60!black}\cmark & 68.4 & 84.2 & 61.2 & 54.7 & 66.8 & 72.0 & 71.3 & 59.6 \\
    & PatchCore    & \color{green!60!black}\cmark & 72.4 & 86.2 & 89.6 & 74.2 & 60.2 & 64.3 & 82.4 & 68.1 \\
    & FastRecon    & \color{green!60!black}\cmark & 51.7 & 74.9 & 62.4 & 59.9 & 58.2 & 74.6 & 79.6 & 63.5 \\
    & WinCLIP      & \color{red!70!black}\xmark & 92.7 & 96.3 & 92.4 & 83.9 & 83.7 & 84.9 & 95.1 & 81.8 \\
    & PromptAD     & \color{red!70!black}\xmark & 89.2 & 94.8 & 92.2 & 84.3 & 84.3 & 87.8 & 96.9 & 84.7 \\
    & AnomalySD    & \color{red!70!black}\xmark & 94.8 & 97.0 & 95.8 & \second{90.4} & \second{87.4} & \second{90.1} & 96.8 & \second{94.1} \\
    & IIPAD        & \color{red!70!black}\xmark & \second{95.7} & \second{97.9} & \second{96.7} & 90.3 & 86.7 & 88.6 & \second{97.2} & 87.9 \\
    & \name{} \textbf{(Ours)}
                   & \color{green!60!black}\cmark & \best96.8 & \best98.3 & \best97.0 & \best93.3 & \best93.5 & \best93.0 & \best99.7 & \best98.0 \\

    \midrule

    %--------- 4-shot + Multi-class ---------
    \multirow{9}{*}{\shortstack{4}}
    & SPADE        & \color{green!60!black}\cmark & 76.6 & 88.8 & 62.8 & 55.6 & 73.0 & 76.6 & 72.1 & 60.9 \\
    & PatchCore    & \color{green!60!black}\cmark & 74.9 & 88.8 & 92.6 & 80.8 & 62.6 & 69.9 & 85.4 & 70.6 \\
    & FastRecon    & \color{green!60!black}\cmark & 50.8 & 73.1 & 65.0 & 62.8 & 57.6 & 73.7 & 78.8 & 62.9 \\
    & WinCLIP      & \color{red!70!black}\xmark & 94.0 & 96.9 & 92.9 & 84.4 & 84.1 & 86.1 & 95.2 & 82.1 \\
    & PromptAD     & \color{red!70!black}\xmark & 90.6 & 96.5 & 92.4 & 84.6 & 85.7 & 88.8 & 97.2 & 84.7 \\
    & AnomalySD    & \color{red!70!black}\xmark & 95.6 & 97.6 & 96.2 & 90.8 & \second{88.9} & \second{90.9} & \second{97.5} & \second{94.3} \\
    & IIPAD        & \color{red!70!black}\xmark & \second{96.1} & \second{98.1} & \second{97.0} & \second{91.2} & 88.3 & 89.6 & 97.4 & 88.3 \\
    & \name{} \textbf{(Ours)}  
                   & \color{green!60!black}\cmark & \best{97.1} & \best{98.6} & \best{97.2} & \best{93.5} & \best94.4 & \best94.0 & \best99.7 & \best98.4 \\
    \bottomrule
    \end{tabular}}
\caption*{\scriptsize The multi-class and few-shot results of SPADE~\citep{spade}, PatchCore~\citep{PatchCore}, FastRecon~\citep{fang2023fastrecon}, WinCLIP, and PromptAD are from IIPAD \citep{lv2025oneforall}.}
\vspace{-5mm}
\end{table}

\paragraph{Quantitative Comparison with Multi-Class Baselines}
As shown in \autoref{table:results_multi_class}, \name{} consistently achieves the best performance across both image-level classification and pixel-level segmentation metrics. Classical few-shot methods such as SPADE, PatchCore, and FastRecon suffer from severe performance degradation when extended to the multi-class setting, highlighting the difficulty of adapting them to various categories. In contrast, \name{} handles multi-class adaptation effectively with its simple projector architecture, even under limited training data. Notably, \name{} achieves strong results \emph{without relying on text prompts}, distinguishing it from recent prompt-based baselines.
On MVTec-AD, \name{} shows a clear advantage under the low-shot regime. For instance, in the 1-shot case, it surpasses the second-best method IIPAD by $1.9\%$ in I-AUROC and $\mathbf{3.0\%}$ in PRO. On VisA, \name{} delivers nearly perfect localization, with pixel-level AUROC reaching $\mathbf{99.7\%}$ consistently across all few-shot settings, outperforming IIPAD by up to $\mathbf{2.8\%}$.
As the number of shots increases to $2$ and $4$, \name{} not only maintains competitive image-level accuracy but also further strengthens its lead in localization metrics. These results highlight the robustness and strong generalization ability of \name{} in the challenging multi-class few-shot anomaly detection scenario.

\begin{table*}[h]
    \centering
    \footnotesize 
    \caption{ \small
    Comparison of results on MVTec-AD and VisA against various \textbf{one-class-one-model} few-shot methods. Ours remains the \textbf{multi-class-one-model} few-shot setting.
    Metrics include image-level AUROC (\%), AUPR (\%), pixel-level P-AUROC (\%), and PRO (\%).  We color the \bestbox{} and the \secondbox{} in the k-shot setting.
    }
    \label{table:results_single_class}
    \resizebox{0.95\textwidth}{!}{%
    \begin{tabular}{llccccccccc}
    \toprule
    \multirow{2}{*}{\textbf{Shot}} & \multirow{2}{*}{\textbf{Method}} & \multirow{2}{*}{\textbf{w/o  Texts}}
      & \multicolumn{4}{c}{\textbf{MVTec-AD}} 
      & \multicolumn{4}{c}{\textbf{VisA}} \\
    \cmidrule(l{1em}r{1em}){4-7}
    \cmidrule(l{1em}r{1em}){8-11}
     &  & & {I-AUROC} & {AUPR} & {P-AUROC} & {PRO}
        & {I-AUROC} & {AUPR} & {P-AUROC} & {PRO} \\
    \midrule

    %---------------- 1-shot ----------------
    \multirow{8}{*}{1}
    & SPADE        & \color{green!60!black}\cmark & 81.0 & 90.6 & 91.2 & 83.9 & 79.5 & 82.0 & 95.6 & 84.1 \\
    & PatchCore    & \color{green!60!black}\cmark & 83.4 & 92.2 & 92.0 & 79.7 & 79.9 & 82.8 & 95.4 & 80.5 \\
    & WinCLIP      & \color{red!70!black}\xmark & 93.1 & 96.5 & 95.2 & 87.1 & 83.8 & 85.1 & 96.4 & {85.1} \\
    & InCTRL$^1$       & \color{red!70!black}\xmark & 91.3 & 95.2 & 94.6 & 87.8 & 83.2 & 84.1 & 89.0 & 66.7\\
    & AnomalyCLIP$^2$  & \color{red!70!black}\xmark & {95.2} & {97.2} & 94.6 & 87.6 & {87.7} & {87.7} & 83.2 & \second{90.1}\\
    & PromptAD     & \color{red!70!black}\xmark & {94.6} & {97.1} & {95.9}& {87.9} & {86.9} & {88.4} & {96.7} & {85.1}\\
    & LogSAD$^3$       & \color{red!70!black}\xmark & \second{95.5} & \second{97.3} & \best{97.0} & \second{92.5} & \second{89.8} & \second{90.3} & \second{97.5} & 88.2 \\
    & \name{} \textbf{(Ours)} 
                   & \color{green!60!black}\cmark & \best96.1 & \best97.9 & \second96.8 & \best92.8 & \best92.6 & \best92.0 & \best99.7 & \best98.0\\

    \midrule

    %---------------- 2-shot ----------------
    \multirow{8}{*}{2}
    & SPADE        & \color{green!60!black}\cmark & 82.9 & 91.7 & 92.0 & 85.7 & 80.7 & 82.3 & 96.2 & 85.7 \\
    & PatchCore    & \color{green!60!black}\cmark & 86.3 & 93.8 & 93.3 & 82.3 & 81.6 & 84.8 & 96.1 & 82.6 \\
    & WinCLIP      & \color{red!70!black}\xmark & 94.4 & 97.0 & 96.0 & 88.4 & 84.6 & 85.8 & 96.8 & {86.2} \\
    & InCTRL$^1$       & \color{red!70!black}\xmark & 91.8 & 95.5 & 95.2 & 88.3 & 86.3 & 86.8 & 89.8 & 68.1\\
    & AnomalyCLIP$^2$  & \color{red!70!black}\xmark & 95.4 & 97.3 & 94.9 & 87.8 & 87.8 & 89.1 & 84.5 & \second{90.8}\\
    & PromptAD     & \color{red!70!black}\xmark & {95.7} & \second{97.9} & {96.2} & {88.5} & {88.3} & {90.0} & {97.1} & 85.8 \\
    & LogSAD$^3$       & \color{red!70!black}\xmark & \second{96.3} & 97.6 & \best{97.3} & \second{93.1} & \second{91.8} & \second{92.0} & \second{97.8} & 89.7 \\
    & \name{} \textbf{(Ours)} 
                   & \color{green!60!black}\cmark & \best96.9 & \best98.3 & \second97.0 & \best93.2 & \best93.8 & \best93.3 & \best99.7 & \best98.2 \\
    \midrule

    %---------------- 4-shot ----------------
    \multirow{8}{*}{4}
    & SPADE        & \color{green!60!black}\cmark & 84.8 & 92.5 & 92.7 & 87.0 & 81.7 & 83.4 & 96.6 & 87.3 \\
    & PatchCore    & \color{green!60!black}\cmark & 88.8 & 94.5 & 94.3 & 84.3 & 85.3 & 87.5 & 96.8 & 84.9 \\
    & WinCLIP      & \color{red!70!black}\xmark & 95.2 & 97.3 & 96.2 & 89.0 & 87.3 & 88.8 & 97.2 & {87.6} \\
    & InCTRL$^1$      & \color{red!70!black}\xmark & 93.1 & 96.3 & 95.8 & 89.5 & 87.8 & 88.0 & 90.2 & 68.8 \\
    & AnomalyCLIP$^2$ & \color{red!70!black}\xmark & 96.1 & 97.8 & 95.5 & 88.2 & 88.8 & 90.1 & 85.2 & \second{91.4} \\
    & PromptAD     & \color{red!70!black}\xmark & \second{96.6} & \second{98.5} & {96.5} & {90.5} & {89.1} & {90.8} & {97.4} & 86.2 \\
    & LogSAD$^3$       & \color{red!70!black}\xmark & \second{96.6} & 97.7 & \best{97.5} & \second{93.5} & \second{93.2} & \second{93.4} & \second{98.1} & 90.5 \\
    & \name{} \textbf{(Ours)}  
                   & \color{green!60!black}\cmark & \best{97.1} & \best{98.6} & \second{97.2} & \best{93.6} & \best94.4 & \best94.0 & \best99.7 & \best98.4 \\
    
    \bottomrule
    \end{tabular}}
    \caption*{\scriptsize $^1$InCTRL~\citep{zhu2024generalistanomalydetectionincontext} only targeted at the image-level anomaly detection, yet we report pixel-level metrics for readers' comprehension. $^2$AnomalyCLIP~\citep{zhouanomalyclip} was originally designed for zero-shot detection. Following AdaptCLIP~\citep{gao2025adaptclip}, we extend it to support detection with few-shot references. $^3$LogSAD originally only provides I-/P-AUROC, and we reproduce the experiment in three rounds and report averaged results on every metric, adhering to our protocols.}
    \vspace{-5mm}
\end{table*}

\paragraph{Quantitative Comparison with One-Class Baselines}
As shown in \autoref{table:results_single_class}, even against few-shot baselines specialized for the one-class-one-model setting, \name{} consistently achieves top performance on both MVTec-AD and VisA. In the 1-shot setting, our method already surpasses strong baselines PromptAD and the previous state-of-the-art LogSAD, with a particularly large margin on VisA by $\mathbf{2.2\%}$ in P-AUROC and $\mathbf{9.8\%}$ in PRO over LogSAD.
In the 4-shot scenario on MVTec-AD, \name{} outperforms PromptAD by $0.7\%$ in P-AUROC and $\mathbf{3.1\%}$ in PRO, while achieving slightly lower P-AUROC than LogSAD. On VisA, however, \name{} surpasses LogSAD by $1.6\%$ in P-AUROC and $\mathbf{7.9\%}$ in PRO, demonstrating consistent superiority in both classification and localization.
Moreover, \name{} relies on a single visual foundation model, which makes it more efficient than LogSAD’s three-model DINO-CLIP-SAM pipeline. Importantly, our improvements are achieved in the more challenging \textit{multi-class-one-model} setting, whereas other baselines operate under the simpler one-class formulation with class-specific memory banks or text prompts. This underscores the robustness and learning ability of \name{}, which remains effective even under harder problem constraints.

\begin{wrapfigure}{r}{0.48\textwidth}

  \captionsetup{type=figure}
  \includegraphics[width=\linewidth]{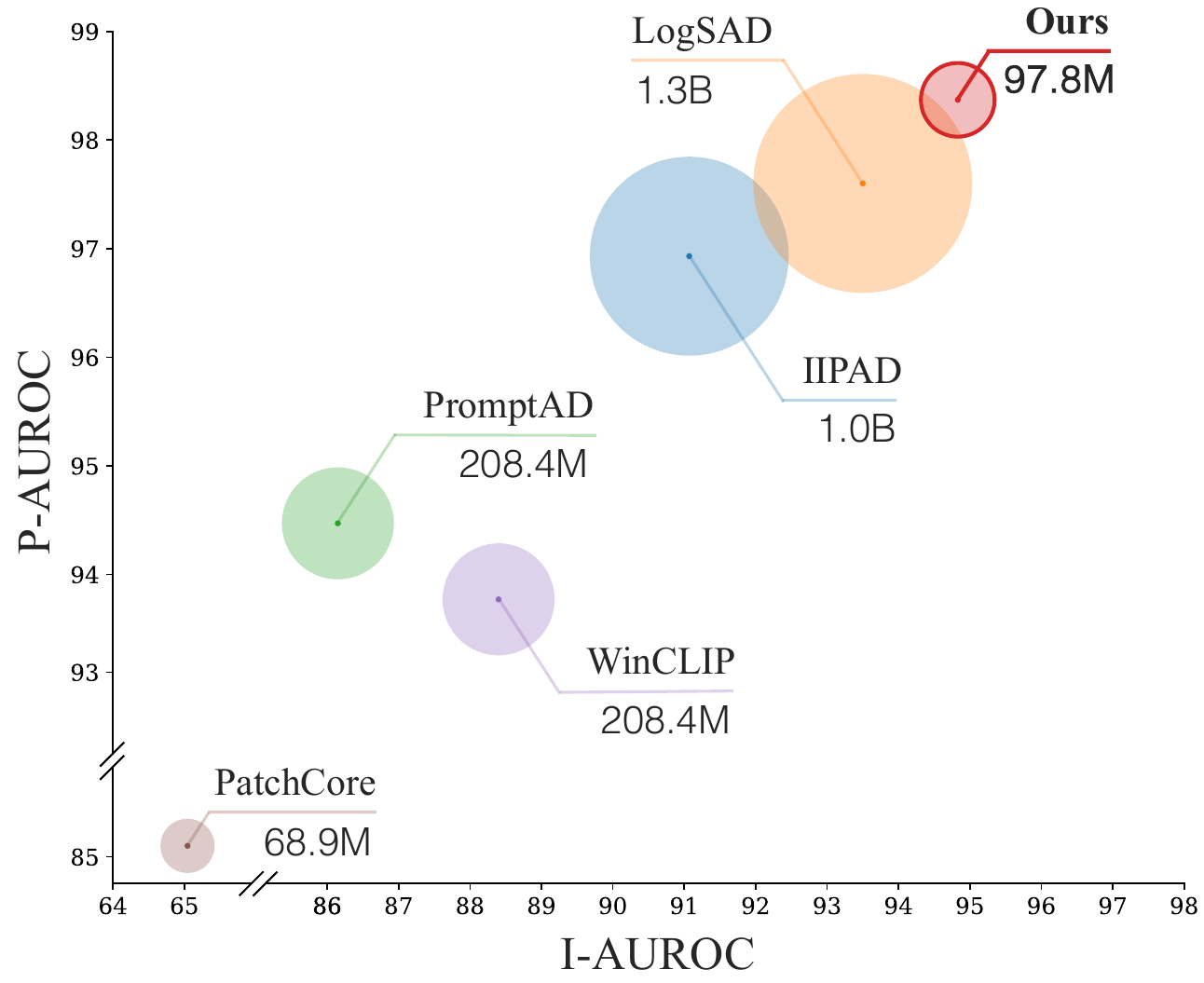}
  \vspace{-1.8em}
  \captionof{figure}{\small  Bubble chart of AUROC results across different methods, averaged over MVTec-AD and VisA from~\autoref{table:results_multi_class}. The smaller the circle is, the fewer parameters it has.}
  \label{params}
  \vspace{-1.5em}
\end{wrapfigure}
\paragraph{Inference Efficiency}
We report the inference time and memory consumption following~\citep{batzner2024efficientad}.
Our projector consists of $11.8$\,M trainable parameters. With DINOv3 as the backbone, \name{} contains $97.8$\,M parameters in total and achieves an average inference time of $128.7$\,ms per image, corresponding to a throughput of approximately $7.8$ images per second, with a peak memory consumption of $1,386$\,MiB on a single RTX~3090. To assess the inference efficiency, we analyze the trade-off between performance and model size across various representative baselines. As shown in \autoref{params}, \name{} achieves overall the best accuracy while using at least one order of magnitude fewer parameters than the large models LogSAD ($\mathbf{\approx 13.3 \times}$) and  IIPAD ($\mathbf{\approx 10.3 \times}$). Despite the compact size, \name{} maintains desirable efficiency, especially suitable in industrial usages.

\begin{figure}[h]
  \centering
  \vspace{14px}
  \begin{overpic}[width=\textwidth,trim=0cm 0 0 0cm,clip]{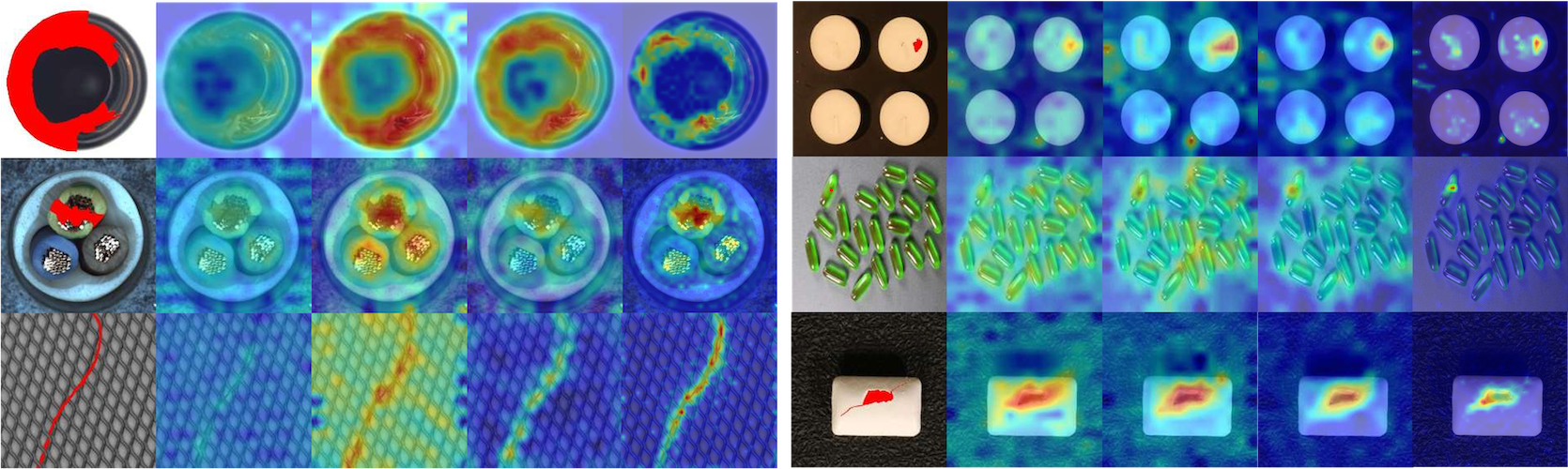}
    \put(19,-2.5){\color{black}{\small (a) MVTec-AD}}
    \put(70.5,-2.5){\color{black}{\small (b) VisA}}

    \put(3.5,31){\color{black}\small {GT}}
    \put(10,31){\color{black}\small {PromptAD}}
    \put(20.5,31){\color{black}\small {WinCLIP}}
    \put(31.8,31){\color{black}\small {IIPAD}}
    \put(42,31){\color{black}\small \textbf{Ours}}
    
    \put(53.85,31){\color{black}\small {GT}}
    \put(60.5,31){\color{black}\small {PromptAD}}
    \put(71,31){\color{black}\small {WinCLIP}}
    \put(82.5,31){\color{black}\small {IIPAD}}
    \put(92.5,31){\color{black}\small \textbf{Ours}}
    
    \put(-2.7,22){\rotatebox{90}{\small Bottle}}
    \put(-2.7,12){\rotatebox{90}{\small Cable}}
    \put(-2.7,3){\rotatebox{90}{\small Grid}}

    \put(101.2,28){\rotatebox{-90}{\small Candle}}
    \put(100.7,19){\rotatebox{-90}{\small Capsules}}
    \put(101.2,7.5){\rotatebox{-90}{\small Gum}}

  \end{overpic}
  \vspace{0.5px}
    \caption{\small \textbf{Qualitative comparison with few-shot baselines in 1-shot setting.} We directly compare our results with the ones cropped from IIPAD~\citep{lv2025oneforall}.
}
    \label{fig:qualitative_comparison}
    \vspace{-3mm}
\end{figure}
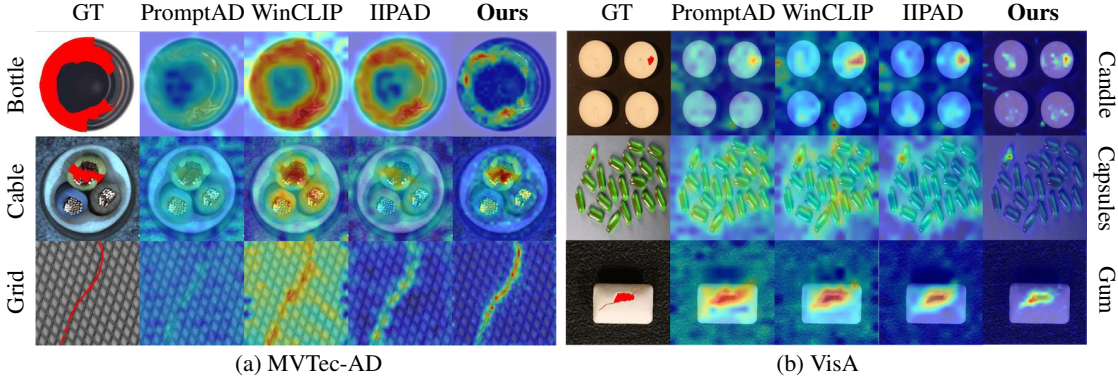

\paragraph{Qualitative Comparison}
We present qualitative comparisons in \autoref{fig:qualitative_comparison}. \name{} effectively localizes anomalous regions with high precision on both datasets. It precisely localizes anomalous regions, successfully capturing both large structural defects and subtle, fine-grained anomalies. In contrast to other methods, \name{} yields cleaner segmentations with substantially less noise in background areas. Additional examples are provided in Supplementary Material~\ref{add:logsad} and~\ref{more_qualitative_results}.

\subsection{Ablation Study}

\begin{table}[h]
    \centering
    \footnotesize
    \caption{\small Comparison of performance with different encoders as the backbones in a 1-shot setting on MVTec-AD. We color the \bestbox{} and the \secondbox{}.}
    \resizebox{0.95\textwidth}{!}{
    \begin{tabular}{lccccccc}
        \toprule
         & DINOv3 & DINOv2 & DINOSigLIP & DINO & SigLIP & CLIP & WideResnet \\
        \midrule
        Pre-trained w/o Texts & \color{green!60!black}\cmark & \color{green!60!black}\cmark & \color{red!70!black}\xmark & \color{green!60!black}\cmark & \color{red!70!black}\xmark & \color{red!70!black}\xmark & \color{green!60!black}\cmark \\
        \midrule
        I-AUROC & \best{96.1} & \second{95.2} & {92.5} & 88.3 & 87.8 & 79.0 & 73.1 \\
        AUPR    & \best{97.9} & \second{97.4} & {95.1} & 94.2 & 93.8 & 87.9 & 87.2 \\
        P-AUROC & \best{96.8} & \second{96.4} & 93.1 & {96.2} & 86.0 & 90.9 & 89.4 \\
        PRO     & \best{92.8} & \second{92.5} & 87.2 & {87.8} & 71.1 & 70.9 & 75.6 \\
        \bottomrule
    \end{tabular}
    }
    \label{ablation_encoder}
\end{table}
\paragraph{Comparison of Different Foundation Models}
To further investigate the capability of different visual encoders on \name{}, we conduct an ablation study comparing several commonly used encoders in the 1-shot setting on MVTec-AD, as shown in \autoref{ablation_encoder}. For VLMs, we utilize only the vision backbone to extract features in the latent space. Our results demonstrate that DINOv3 achieves the best overall performance. Among the alternative encoders, the DINO series remains competitive, with DINOv2 and DINOSigLIP reaching I-AUROC scores of $95.2\%$ and $92.5\%$, respectively. As expected, CLIP lags markedly, achieving only $70.9\%$ PRO, consistent with WinCLIP’s observation that CLIP lacks pixel-level information~\citep{jeong2023winclip}. This also suggests that CLIP is less effective than SigLIP in learning natural image manifold for fine-grained anomaly localization in few-shot settings. WideResNet shows the weakest performance, indicating that traditional CNNs, even pretrained, struggle to generalize well under minimal supervision.

Most importantly, \autoref{ablation_encoder} reveals that foundation features pre-trained from pure visual supervision, without alignment on textual information, can still deliver highly competitive performance for anomaly detection in challenging few-shot settings. This finding highlights that textual features are not a necessity; instead, the representational power of strong visual features alone can be fully leveraged to uncover anomalies with minimal data.

\begin{wrapfigure}{r}{0.55\textwidth}
  \vspace{-1.5em}
  \captionsetup{type=figure}
  \includegraphics[width=\linewidth]{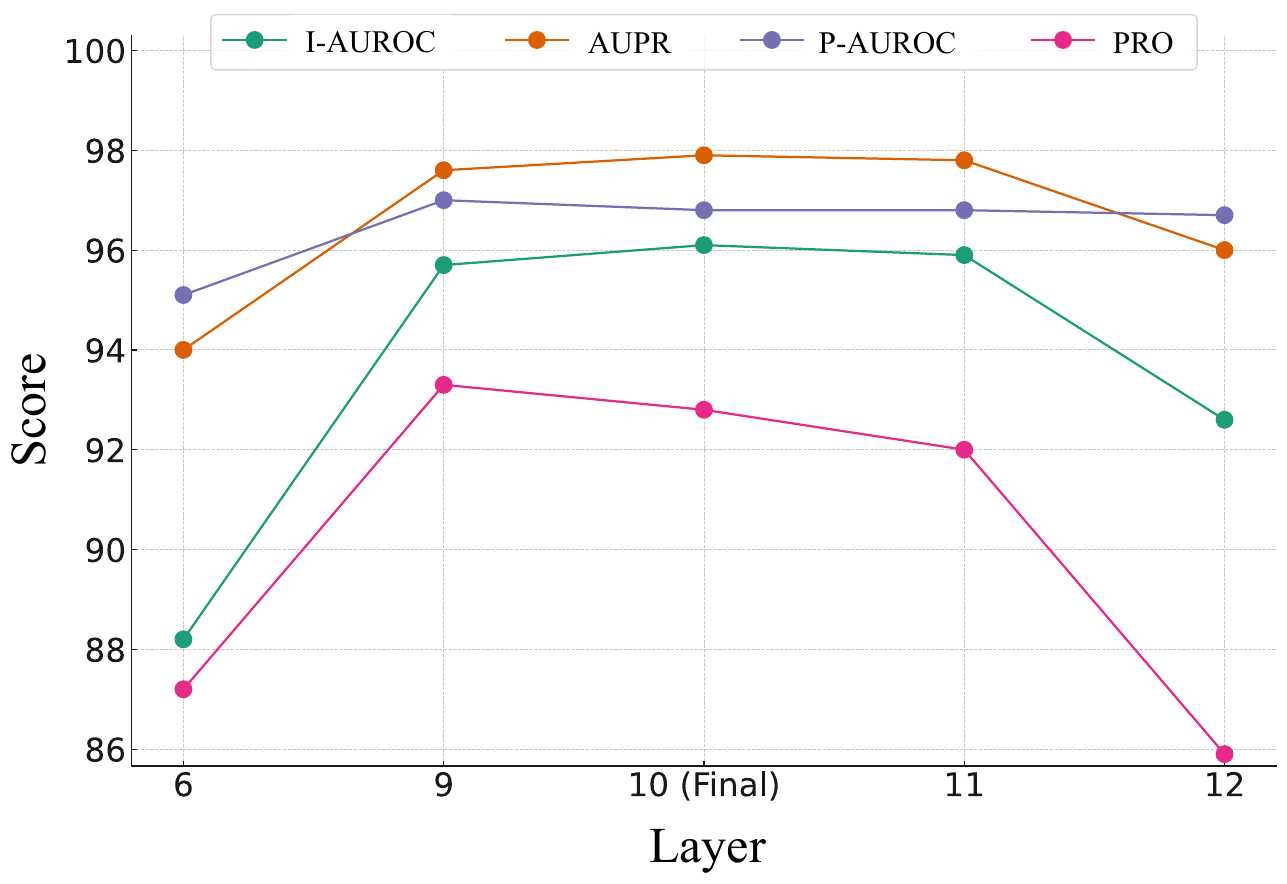}
  \vspace{-1.8em}
  \captionof{figure}{\small  Comparison of performance using different layers of DINOv3 in a 1-shot setting on MVTec-AD.}
  \label{ablation_layer}
  \vspace{-1em}
\end{wrapfigure}
\paragraph{Comparison between DINOv3 Layers}
We analyze the effect of selecting different layers in DINOv3 for feature extraction in the 1-shot setting on MVTec-AD, as shown in \autoref{ablation_layer}. Layer 10, used in \name{}, achieves the overall best performance, with the highest I-AUROC ($96.1\%$), AUPR ($97.9\%$), and P-AUROC ($96.8\%$), and a competitive PRO ($92.8\%$). While adjacent layers such as layer 9 and 11 yield comparable results, performance declines when using shallower (e.g., layer 6) or deeper layers (e.g., layer 12), especially in terms of I-AUROC and PRO. This indicates that mid-to-late layers strike a better balance between semantics and spatial precision, while overly abstract or low-level features are suboptimal for fine-grained localization.

\begin{wraptable}{r}{0.55\textwidth} % r=右侧, l=左侧
\centering
\vspace{-1.2em}
\caption{\small 1-shot performance comparison of different architectures of the manifold projector on MVTec-AD.}
\label{tab:Different_projector}
\resizebox{\linewidth}{!}{
\begin{tabular}{lccccc}
\toprule
\textbf{Type} & \textbf{Depth} & \textbf{I-AUROC} & \textbf{AUPR} & \textbf{P-AUROC} & \textbf{PRO} \\ \midrule
\multirow{3}{*}{{ViT}} & 4 & 95.5 & 97.2 & 96.6 & 92.6 \\
                             & 6 & {96.1} & {97.9} & {96.8} & {92.8} \\
                             & 8 & 95.8 & 97.3 & 96.8 & 92.5 \\ \midrule
\multirow{3}{*}{{MLP}} & 4 & 93.5 & 96.2 & 95.7 & 91.2 \\
                             & 6 & 92.1 & 95.4 & 95.2 & 90.7 \\
                             & 8 & 87.8 & 91.9 & 91.2 & 82.1 \\ \bottomrule
\end{tabular}
}
\vspace{-0.6em}
\end{wraptable}
\paragraph{Comparison of Projector Designs} In our initial exploration of projector architectures, we compared two widely used modules, MLP layers and ViT attention blocks, under different layer configurations in the 1-shot setting on MVTec-AD. As shown in~\autoref{tab:Different_projector}, ViT consistently outperforms MLP with the same depth. This advantage arises from the self-attention mechanism, which enables richer patch-wise interactions and improves the detection of fine-grained anomalies. However, simply increasing the network depth does not necessarily yield further gains and instead adds computational overhead. Guided by these observations, we adopt a 6-layer self-attention block in our final design, as it offers a favorable trade-off between accuracy and computational cost.

\paragraph{Top-K Selection}
The optimal $K$ varies across datasets, as the size and distribution of abnormal regions differ among various datasets \citep{strater2024generalad}. We evaluate the effect of different values of $K$ in the Top-K selection mechanism on MVTec-AD and VisA in a 1-shot setting. \autoref{ablation_top_k} indicates that increasing $K$ initially improves performance, with the best results observed at $K=10$ for MVTec-AD and $K=6$ for VisA. Beyond these points, performance gradually declines.
\begin{table}[h]
    \centering
    \footnotesize
    \caption{\small Comparison of performance with DINOv3 backbone and different $K$ of Top-$K$ in a 1-shot setting on MVTec-AD and VisA. We color the \bestbox{}.}
    \resizebox{0.9\textwidth}{!}{
    \begin{tabular}{llcccccccc}
        \toprule
        \multirow{2}{*}{{Dataset}}
        & \multirow{2}{*}{{Metric}}
        & \multicolumn{8}{c}{{K}} \\
        \cmidrule(l{1em}r{1em}){3-10}
        & & {1} & {2} & {4} & {6} & {10} & {14} & {16} & {20} \\
        \midrule
        \multirow{2}{*}{MVTec-AD}
          & I-AUROC  & 94.82 & 95.41 & 95.89 & 95.96 & \best{96.09} & 95.91 & 95.86 & 95.74 \\
          & AUPR     & 97.10 & 97.47 & 97.78 & 97.80 & \best{97.92} & 97.81 & 97.78 & 97.75 \\
        \midrule
        \multirow{2}{*}{VisA}
          & I-AUROC  & 91.64 & 92.26 & 92.63 & \best{92.64}  & 92.44 & 92.28 & 92.17 & 91.96 \\
          & AUPR    & 90.91 & 91.39 & 91.56 & \best{91.98} & 91.56 & 91.55 & 91.53 & 91.41 \\
        \bottomrule
    \end{tabular}
    }
    \label{ablation_top_k}
\end{table}
\subsection{Failure Cases}
\label{failure}
\begin{figure}[h]
  \centering
  \vspace{10px}
  \begin{overpic}[width=0.95\textwidth,trim=0cm 0 0 0cm,clip]{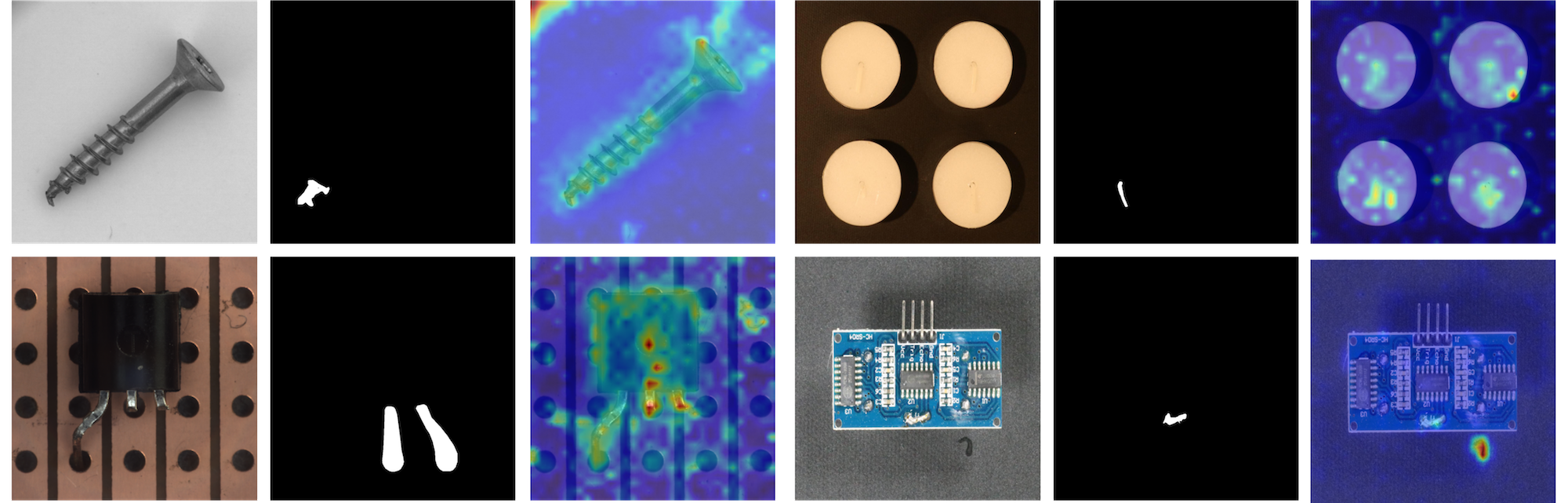}
    \put(17,-3){\color{black}{\small (a) MVTec-AD}}
    \put(70.5,-3){\color{black}{\small (b) VisA}}

    \put(4,33.5){\color{black}\small {Abnormal}}
    \put(18.3,33.5){\color{black}\small {Ground Truth}}
    \put(36,33.5){\color{black}\small {\name{}}}
    \put(54,33.5){\color{black}\small {Abnormal}}
    \put(68.5,33.5){\color{black}\small {Ground Truth}}
    \put(86.2,33.5){\color{black}\small {\name{}}}

    \put(-2.7,21){\rotatebox{90}{\small Screw}}
    \put(-2.7,3){\rotatebox{90}{\small Transistor}}

    \put(100.7,28){\rotatebox{-90}{\small Candle}}
    \put(100.7,11){\rotatebox{-90}{\small Pcb2}}

  \end{overpic}
  \vspace{10px}
    \caption{\small Typical failure cases of \name{}.}
    \label{fig:failure_case}
    \vspace{-1mm}
\end{figure}
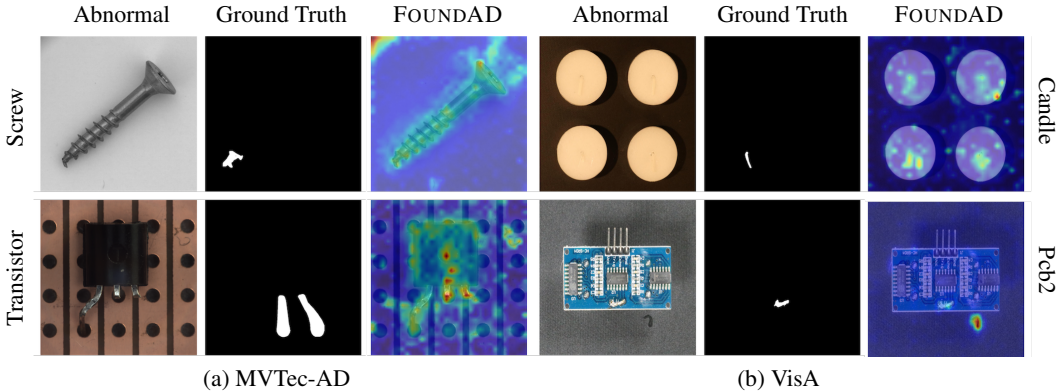

While \name{} achieves generally strong performance in both image classification and anomaly localization, it can still be challenged under certain conditions. Typical failure cases are shown in~\autoref{fig:failure_case}. In~\autoref{fig:failure_case} (a), the test screw appears in a different orientation from the training samples. Without a spatial transformation mechanism to align features as adopted in RegAD~\citep{huang2022registration}, \name{} fails to recognize the anomaly. Similarly, a transistor with missing pins can be localized, but precise segmentation is unsuccessful. We attribute this to two factors: (i) the projector is trained solely on normal data and therefore lacks priors about the morphology of potential anomalies; and (ii) the projector, inspired by I-JEPA~\citep{ijepa}, is non-generative, limiting its ability to complete anomaly masks. In~\autoref{fig:failure_case} (b), the candle exhibits severe appearance variations caused by exposure. The anomaly becomes nearly indistinguishable from the background, hindering representation learning and degrading detection quality. Moreover, \name{} occasionally misclassifies background artifacts as anomalies. For example, previously unseen stains on the platform of Pcb2 are incorrectly highlighted during inference.

% \newpage

\section{Related Work}
\label{sec:formatting}

\paragraph{Multi-class Anomaly Detection}
 Many existing methods follow a one-class-one-model \citep{patchsvdd, spade, defard2021padim, PatchCore, differnet, zavrtanik2021draem, zavrtanik2022dsr, RD, RD++, MSFR, zhang2024realnet, CutPaste, hu2024anomalydiffusion, destseg} paradigm, necessitating a separate model for each class, which limits generalization to unseen categories. To overcome this limitation, general models \citep{huang2022registration, CAReg, mambaad} and multi-class-one-model approaches \citep{you2022unified, zhao2023omnial, lu2023hierarchical, he2024diffusion} have been proposed. In this work, we focus primarily on multi-class settings. UniAD \citep{you2022unified} presents a unified framework capable of handling multiple classes simultaneously, while HVQ-Trans \citep{lu2023hierarchical} employs vector quantization to learn a structured latent space that enhances anomaly detection within each category and preserves cross-category discrimination. DiAD \citep{zhang2023diffusionad} introduces a diffusion-based method for multi-class anomaly detection. Notably, these methods typically require a large amount of training data, which is impractical in real industrial scenarios since collecting and labeling samples is challenging and expensive. Therefore, we aim to develop an approach for multi-class anomaly detection with minimal training effort.

\paragraph{Few-shot Anomaly Detection}Few-shot anomaly detection aims to detect and localize anomalies using minimal training samples per category \citep{fang2023fastrecon, duan2023few, damm2024anomalydinoboostingpatchbasedfewshot, zhu2024generalistanomalydetectionincontext, huang2022registration, CAReg, jeong2023winclip, anomalygpt, yan2024anomalysd, lv2025oneforall}. For example, RegAD \citep{huang2022registration} introduces spatial transformation consistency to ensure robustness under few-shot conditions, while FastRecon \citep{fang2023fastrecon} leverages an efficient feature reconstruction approach for rapid adaptation to unseen anomaly types. Another line of research exploits textual information to support few-shot anomaly detection; methods such as AnomalyGPT \citep{anomalygpt}, InCTRL \citep{zhu2024generalistanomalydetectionincontext}, WinCLIP \citep{jeong2023winclip}, and PromptAD \citep{li2024promptad} employ vision-language models to address the few-shot detection problem. Moreover, recent work on the multi-class setting, such as AnomalySD \citep{yan2024anomalysd} and IIPAD \citep{lv2025oneforall}, has proposed unified models that handle multiple categories within a single framework while achieving competitive performance. 
LogSAD \citep{zhang2025towards} leverages GPT-4V~\citep{achiam2023gpt} to generate textual reasoning rules per category and combines patch-level and set-level detections with score calibration, achieving strong performance in a training-free manner.
However, relying on textual inputs or support images during inference remains impractical in many real-world applications.

\paragraph{Foundation Models for Visual Understanding}  
Foundation models have driven major advances in visual understanding~\citep{zhou2022extract,gu2024conceptgraphs,liu2024grounding}, content generation~\citep{zhai2024echoscene,zhai2023commonscenes,karras2023dreampose}, vision-based planning~\citep{zhou2024dino,kim2024openvla}, geometry estimation~\citep{chen2024secondpose,wang2025vggt,edstedt2024roma}, and future prediction~\citep{karypidis2024dino,baldassarre2025back}. Early supervised pretraining on large-scale datasets such as ImageNet~\citep{deng2009imagenet} with convolutional networks~\citep{he2016deep,zagoruyko2016wide} established strong transferable features, while self-supervised contrastive learning, such as SimCLR~\citep{chen2020simple}, MoCo~\citep{he2020momentum}, further improved label-free representation learning. Vision Transformers, particularly the DINO series~\citep{caron2021emerging,Dinov2,simeoni2025dinov3}, enriched semantic capture, and masked image modeling~\citep{mae,bao2021beit,ijepa} showed that predicting masked tokens yields highly transferable embeddings. Vision-language models such as CLIP~\citep{radford2021learning} and SigLIP~\citep{zhai2023sigmoid} extend this by grounding visual features in text.  
These pre-trained representations form well-structured manifolds that benefit anomaly detection, where distinguishing normal from abnormal patterns hinges on the underlying feature space~\citep{heckler2023exploring}. Inspired particularly by I-JEPA~\citep{ijepa}, which predicts masked tokens directly in latent space, we build on strong embedding structures and introduce a projector to map mixed anomalous and normal feature tokens back to the normal manifold.  

\section{Conclusion}
\label{sec:con}

In this work, we presented a novel few-shot anomaly detection approach, \name{}, leveraging the pure visual embeddings from foundation encoders \emph{without text-prompt assistance}. By uncovering the direct correlation between embedding distance and anomaly amount, we designed a lightweight nonlinear feature projector that efficiently maps features onto the normal image manifold. This method effectively addresses the challenge posed by limited anomaly examples, achieving robust detection with minimal training data. Extensive evaluations across multiple foundational visual encoders demonstrated that our method surpasses current few-shot anomaly detection methods. Our findings emphasize the potential of foundation visual encoders solely for anomaly detection tasks, advocating for broader exploration and application in industrial inspection scenarios.

\newpage
% \paragraph{Reproducibility Statement}
% We have made extensive efforts to ensure the reproducibility of our work. The architecture details of our framework, including the encoders and the non-linear projector, are described in~\autoref{sec:AD-JEPA}. Training settings, hyperparameters, and evaluation protocols are provided in~\autoref{sec:exp} and Supplementary Material~\ref{encoder_ablation_detail}. All datasets used (MVTec-AD and VisA) are publicly available. Additional qualitative and quantitative results are included in Supplementary Material~\ref{additional}. Finally, we provide our code and instructions at { \small \url{ https://github.com/ymxlzgy/FoundAD}}.

% \paragraph{LLM Usage}
% We used ChatGPT solely as a writing assistant for grammar checking and language polishing of the manuscript. All research ideas, experiments, analyses, and conclusions were conceived and conducted entirely by the authors. ChatGPT and other LLMs were not used for obtaining ideas, for technical content generation, or for experimental design.

{
    \small
    \bibliographystyle{gy_template}
    \bibliography{main}

\begin{thebibliography}{70}
\providecommand{\natexlab}[1]{#1}
\providecommand{\url}[1]{\texttt{#1}}
\expandafter\ifx\csname urlstyle\endcsname\relax
  \providecommand{\doi}[1]{doi: #1}\else
  \providecommand{\doi}{doi: \begingroup \urlstyle{rm}\Url}\fi

\bibitem[Achiam et~al.(2023)Achiam, Adler, Agarwal, Ahmad, Akkaya, Aleman, Almeida, Altenschmidt, Altman, Anadkat, et~al.]{achiam2023gpt}
Josh Achiam, Steven Adler, Sandhini Agarwal, Lama Ahmad, Ilge Akkaya, Florencia~Leoni Aleman, Diogo Almeida, Janko Altenschmidt, Sam Altman, Shyamal Anadkat, et~al.
\newblock Gpt-4 technical report, 2023.
\newblock URL \url{https://arxiv.org/abs/2303.08774}.

\bibitem[Assran et~al.(2023)Assran, Duval, Misra, Bojanowski, Vincent, Rabbat, LeCun, and Ballas]{ijepa}
Mahmoud Assran, Quentin Duval, Ishan Misra, Piotr Bojanowski, Pascal Vincent, Michael Rabbat, Yann LeCun, and Nicolas Ballas.
\newblock Self-supervised learning from images with a joint-embedding predictive architecture.
\newblock In \emph{CVPR}, 2023.

\bibitem[Baldassarre et~al.(2025)Baldassarre, Szafraniec, Terver, Khalidov, Massa, LeCun, Labatut, Seitzer, and Bojanowski]{baldassarre2025back}
Federico Baldassarre, Marc Szafraniec, Basile Terver, Vasil Khalidov, Francisco Massa, Yann LeCun, Patrick Labatut, Maximilian Seitzer, and Piotr Bojanowski.
\newblock Back to the features: Dino as a foundation for video world models, 2025.
\newblock URL \url{https://arxiv.org/abs/2507.19468}.

\bibitem[Bao et~al.(2022)Bao, Dong, Piao, and Wei]{bao2021beit}
Hangbo Bao, Li~Dong, Songhao Piao, and Furu Wei.
\newblock Beit: Bert pre-training of image transformers.
\newblock In \emph{ICLR}, 2022.

\bibitem[Batzner et~al.(2024)Batzner, Heckler, and K{\"o}nig]{batzner2024efficientad}
Kilian Batzner, Lars Heckler, and Rebecca K{\"o}nig.
\newblock Efficientad: Accurate visual anomaly detection at millisecond-level latencies.
\newblock In \emph{CVPR}, 2024.

\bibitem[Bergmann et~al.(2019)Bergmann, Fauser, Sattlegger, and Steger]{bergmann2019mvtec}
Paul Bergmann, Michael Fauser, David Sattlegger, and Carsten Steger.
\newblock Mvtec ad--a comprehensive real-world dataset for unsupervised anomaly detection.
\newblock In \emph{CVPR}, 2019.

\bibitem[Caron et~al.(2021)Caron, Touvron, Misra, J{\'e}gou, Mairal, Bojanowski, and Joulin]{caron2021emerging}
Mathilde Caron, Hugo Touvron, Ishan Misra, Herv{\'e} J{\'e}gou, Julien Mairal, Piotr Bojanowski, and Armand Joulin.
\newblock Emerging properties in self-supervised vision transformers.
\newblock In \emph{CVPR}, 2021.

\bibitem[Chen et~al.(2020)Chen, Kornblith, Norouzi, and Hinton]{chen2020simple}
Ting Chen, Simon Kornblith, Mohammad Norouzi, and Geoffrey Hinton.
\newblock A simple framework for contrastive learning of visual representations.
\newblock In \emph{ICML}, 2020.

\bibitem[Chen \& He(2021)Chen and He]{chen2021exploring}
Xinlei Chen and Kaiming He.
\newblock Exploring simple siamese representation learning.
\newblock In \emph{CVPR}, 2021.

\bibitem[Chen et~al.(2024)Chen, Di, Zhai, Manhardt, Zhang, Zhang, Tombari, Navab, and Busam]{chen2024secondpose}
Yamei Chen, Yan Di, Guangyao Zhai, Fabian Manhardt, Chenyangguang Zhang, Ruida Zhang, Federico Tombari, Nassir Navab, and Benjamin Busam.
\newblock Secondpose: Se (3)-consistent dual-stream feature fusion for category-level pose estimation.
\newblock In \emph{CVPR}, 2024.

\bibitem[Cohen \& Hoshen(2020)Cohen and Hoshen]{spade}
N~Cohen and Y~Hoshen.
\newblock Sub-image anomaly detection with deep pyramid correspondences, 2020.
\newblock URL \url{https://arxiv.org/abs/2005.02357}.

\bibitem[Damm et~al.(2024)Damm, Laszkiewicz, Lederer, and Fischer]{damm2024anomalydinoboostingpatchbasedfewshot}
Simon Damm, Mike Laszkiewicz, Johannes Lederer, and Asja Fischer.
\newblock Anomalydino: Boosting patch-based few-shot anomaly detection with dinov2, 2024.
\newblock URL \url{https://arxiv.org/abs/2405.14529}.

\bibitem[Defard et~al.(2021)Defard, Setkov, Loesch, and Audigier]{defard2021padim}
Thomas Defard, Aleksandr Setkov, Angelique Loesch, and Romaric Audigier.
\newblock Padim: a patch distribution modeling framework for anomaly detection and localization.
\newblock In \emph{ICPR}, 2021.

\bibitem[Deng \& Li(2022)Deng and Li]{RD}
Hanqiu Deng and Xingyu Li.
\newblock Anomaly detection via reverse distillation from one-class embedding.
\newblock In \emph{CVPR}, 2022.

\bibitem[Deng et~al.(2009)Deng, Dong, Socher, Li, Li, and Fei-Fei]{deng2009imagenet}
Jia Deng, Wei Dong, Richard Socher, Li-Jia Li, Kai Li, and Li~Fei-Fei.
\newblock Imagenet: A large-scale hierarchical image database.
\newblock In \emph{CVPR}, 2009.

\bibitem[Duan et~al.(2023)Duan, Hong, Niu, and Zhang]{duan2023few}
Yuxuan Duan, Yan Hong, Li~Niu, and Liqing Zhang.
\newblock Few-shot defect image generation via defect-aware feature manipulation.
\newblock In \emph{AAAI}, 2023.

\bibitem[Edstedt et~al.(2024)Edstedt, Sun, B{\"o}kman, Wadenb{\"a}ck, and Felsberg]{edstedt2024roma}
Johan Edstedt, Qiyu Sun, Georg B{\"o}kman, M{\aa}rten Wadenb{\"a}ck, and Michael Felsberg.
\newblock Roma: Robust dense feature matching.
\newblock In \emph{CVPR}, 2024.

\bibitem[Fang et~al.(2023)Fang, Wang, Li, Liu, Hu, and Xiao]{fang2023fastrecon}
Zheng Fang, Xiaoyang Wang, Haocheng Li, Jiejie Liu, Qiugui Hu, and Jimin Xiao.
\newblock Fastrecon: Few-shot industrial anomaly detection via fast feature reconstruction.
\newblock In \emph{ICCV}, 2023.

\bibitem[Gao et~al.(2025)Gao, Zhou, Yan, Cai, Zhang, Wang, Liu, Liu, Wang, and Wang]{gao2025adaptclip}
Bin-Bin Gao, Yue Zhou, Jiangtao Yan, Yuezhi Cai, Weixi Zhang, Meng Wang, Jun Liu, Yong Liu, Lei Wang, and Chengjie Wang.
\newblock Adaptclip: Adapting clip for universal visual anomaly detection, 2025.
\newblock URL \url{https://arxiv.org/abs/2505.09926}.

\bibitem[Gu et~al.(2024{\natexlab{a}})Gu, Kuwajerwala, Morin, Jatavallabhula, Sen, Agarwal, Rivera, Paul, Ellis, Chellappa, et~al.]{gu2024conceptgraphs}
Qiao Gu, Ali Kuwajerwala, Sacha Morin, Krishna~Murthy Jatavallabhula, Bipasha Sen, Aditya Agarwal, Corban Rivera, William Paul, Kirsty Ellis, Rama Chellappa, et~al.
\newblock Conceptgraphs: Open-vocabulary 3d scene graphs for perception and planning.
\newblock In \emph{ICRA}, 2024{\natexlab{a}}.

\bibitem[Gu et~al.(2024{\natexlab{b}})Gu, Zhu, Zhu, Chen, Tang, and Wang]{anomalygpt}
Zhaopeng Gu, Bingke Zhu, Guibo Zhu, Yingying Chen, Ming Tang, and Jinqiao Wang.
\newblock Anomalygpt: Detecting industrial anomalies using large vision-language models.
\newblock In \emph{AAAI}, 2024{\natexlab{b}}.

\bibitem[He et~al.(2024{\natexlab{a}})He, Bai, Zhang, He, Chen, Gan, Wang, Li, Tian, and Xie]{mambaad}
Haoyang He, Yuhu Bai, Jiangning Zhang, Qingdong He, Hongxu Chen, Zhenye Gan, Chengjie Wang, Xiangtai Li, Guanzhong Tian, and Lei Xie.
\newblock Mambaad: Exploring state space models for multi-class unsupervised anomaly detection.
\newblock In \emph{NeurIPS}, 2024{\natexlab{a}}.

\bibitem[He et~al.(2024{\natexlab{b}})He, Zhang, Chen, Chen, Li, Chen, Wang, Wang, and Xie]{he2024diffusion}
Haoyang He, Jiangning Zhang, Hongxu Chen, Xuhai Chen, Zhishan Li, Xu~Chen, Yabiao Wang, Chengjie Wang, and Lei Xie.
\newblock A diffusion-based framework for multi-class anomaly detection.
\newblock In \emph{AAAI}, 2024{\natexlab{b}}.

\bibitem[He et~al.(2016)He, Zhang, Ren, and Sun]{he2016deep}
Kaiming He, Xiangyu Zhang, Shaoqing Ren, and Jian Sun.
\newblock Deep residual learning for image recognition.
\newblock In \emph{CVPR}, 2016.

\bibitem[He et~al.(2020)He, Fan, Wu, Xie, and Girshick]{he2020momentum}
Kaiming He, Haoqi Fan, Yuxin Wu, Saining Xie, and Ross Girshick.
\newblock Momentum contrast for unsupervised visual representation learning.
\newblock In \emph{CVPR}, 2020.

\bibitem[He et~al.(2022)He, Chen, Xie, Li, Doll{\'a}r, and Girshick]{mae}
Kaiming He, Xinlei Chen, Saining Xie, Yanghao Li, Piotr Doll{\'a}r, and Ross Girshick.
\newblock Masked autoencoders are scalable vision learners.
\newblock In \emph{CVPR}, 2022.

\bibitem[Heckler et~al.(2023)Heckler, K{\"o}nig, and Bergmann]{heckler2023exploring}
Lars Heckler, Rebecca K{\"o}nig, and Paul Bergmann.
\newblock Exploring the importance of pretrained feature extractors for unsupervised anomaly detection and localization.
\newblock In \emph{CVPR}, 2023.

\bibitem[Hu et~al.(2024)Hu, Zhang, Yi, Du, Chen, Liu, Wang, and Wang]{hu2024anomalydiffusion}
Teng Hu, Jiangning Zhang, Ran Yi, Yuzhen Du, Xu~Chen, Liang Liu, Yabiao Wang, and Chengjie Wang.
\newblock Anomalydiffusion: Few-shot anomaly image generation with diffusion model.
\newblock In \emph{AAAI}, 2024.

\bibitem[Huang et~al.(2022)Huang, Guan, Jiang, Zhang, Spratling, and Wang]{huang2022registration}
Chaoqin Huang, Haoyan Guan, Aofan Jiang, Ya~Zhang, Michael Spratling, and Yan-Feng Wang.
\newblock Registration based few-shot anomaly detection.
\newblock In \emph{ECCV}, 2022.

\bibitem[Huang et~al.(2024)Huang, Guan, Jiang, Zhang, Spratling, Wang, and Wang]{CAReg}
Chaoqin Huang, Haoyan Guan, Aofan Jiang, Ya~Zhang, Michael Spratling, Xinchao Wang, and Yanfeng Wang.
\newblock Few-shot anomaly detection via category-agnostic registration learning.
\newblock \emph{IEEE Transactions on Neural Networks and Learning Systems}, 2024.

\bibitem[Iqbal et~al.(2024)Iqbal, Khan, Javed, Moyo, Zweiri, and Abdulrahman]{MSFR}
Ehtesham Iqbal, Samee~Ullah Khan, Sajid Javed, Brain Moyo, Yahya Zweiri, and Yusra Abdulrahman.
\newblock Multi-scale feature reconstruction network for industrial anomaly detection.
\newblock \emph{Knowledge-Based Systems}, 305:\penalty0 112650, 2024.

\bibitem[Jeong et~al.(2023)Jeong, Zou, Kim, Zhang, Ravichandran, and Dabeer]{jeong2023winclip}
Jongheon Jeong, Yang Zou, Taewan Kim, Dongqing Zhang, Avinash Ravichandran, and Onkar Dabeer.
\newblock Winclip: Zero-/few-shot anomaly classification and segmentation.
\newblock In \emph{CVPR}, 2023.

\bibitem[Karamcheti et~al.(2024)Karamcheti, Nair, Balakrishna, Liang, Kollar, and Sadigh]{karamcheti2024prismatic}
Siddharth Karamcheti, Suraj Nair, Ashwin Balakrishna, Percy Liang, Thomas Kollar, and Dorsa Sadigh.
\newblock Prismatic vlms: Investigating the design space of visually-conditioned language models.
\newblock In \emph{ICML}, 2024.

\bibitem[Karras et~al.(2023)Karras, Holynski, Wang, and Kemelmacher-Shlizerman]{karras2023dreampose}
Johanna Karras, Aleksander Holynski, Ting-Chun Wang, and Ira Kemelmacher-Shlizerman.
\newblock Dreampose: Fashion image-to-video synthesis via stable diffusion.
\newblock In \emph{ICCV}, 2023.

\bibitem[Karypidis et~al.(2025)Karypidis, Kakogeorgiou, Gidaris, and Komodakis]{karypidis2024dino}
Efstathios Karypidis, Ioannis Kakogeorgiou, Spyros Gidaris, and Nikos Komodakis.
\newblock Dino-foresight: Looking into the future with dino.
\newblock In \emph{NeurIPS}, 2025.

\bibitem[Kim et~al.(2025)Kim, Pertsch, Karamcheti, Xiao, Balakrishna, Nair, Rafailov, Foster, Lam, Sanketi, et~al.]{kim2024openvla}
Moo~Jin Kim, Karl Pertsch, Siddharth Karamcheti, Ted Xiao, Ashwin Balakrishna, Suraj Nair, Rafael Rafailov, Ethan Foster, Grace Lam, Pannag Sanketi, et~al.
\newblock Openvla: An open-source vision-language-action model.
\newblock \emph{CoRL}, 2025.

\bibitem[LeCun(2022)]{jepa}
Yann LeCun.
\newblock A path towards autonomous machine intelligence version 0.9. 2, 2022-06-27, 2022.
\newblock URL \url{https://openreview.net/pdf?id=BZ5a1r-kVsf}.

\bibitem[Li et~al.(2021)Li, Sohn, Yoon, and Pfister]{CutPaste}
Chun-Liang Li, Kihyuk Sohn, Jinsung Yoon, and Tomas Pfister.
\newblock Cutpaste: Self-supervised learning for anomaly detection and localization.
\newblock In \emph{CVPR}, 2021.

\bibitem[Li et~al.(2024)Li, Goodge, Liu, and Foo]{li2024promptad}
Yiting Li, Adam Goodge, Fayao Liu, and Chuan-Sheng Foo.
\newblock Promptad: Zero-shot anomaly detection using text prompts.
\newblock In \emph{CVPR}, 2024.

\bibitem[Liu et~al.(2024)Liu, Zeng, Ren, Li, Zhang, Yang, Jiang, Li, Yang, Su, et~al.]{liu2024grounding}
Shilong Liu, Zhaoyang Zeng, Tianhe Ren, Feng Li, Hao Zhang, Jie Yang, Qing Jiang, Chunyuan Li, Jianwei Yang, Hang Su, et~al.
\newblock Grounding dino: Marrying dino with grounded pre-training for open-set object detection.
\newblock In \emph{ECCV}, 2024.

\bibitem[Lu et~al.(2023)Lu, Wu, Tian, Wang, Chen, Liu, and Hu]{lu2023hierarchical}
Ruiying Lu, YuJie Wu, Long Tian, Dongsheng Wang, Bo~Chen, Xiyang Liu, and Ruimin Hu.
\newblock Hierarchical vector quantized transformer for multi-class unsupervised anomaly detection.
\newblock In \emph{NeurIPS}, 2023.

\bibitem[Lv et~al.(2025)Lv, Su, and Xu]{lv2025oneforall}
Wenxi Lv, Qinliang Su, and Wenchao Xu.
\newblock One-for-all few-shot anomaly detection via instance-induced prompt learning.
\newblock In \emph{ICLR}, 2025.

\bibitem[Oquab et~al.(2024)Oquab, Darcet, Moutakanni, Vo, Szafraniec, Khalidov, Fernandez, HAZIZA, Massa, El-Nouby, Assran, Ballas, Galuba, Howes, Huang, Li, Misra, Rabbat, Sharma, Synnaeve, Xu, Jegou, Mairal, Labatut, Joulin, and Bojanowski]{Dinov2}
Maxime Oquab, Timoth{\'e}e Darcet, Th{\'e}o Moutakanni, Huy~V. Vo, Marc Szafraniec, Vasil Khalidov, Pierre Fernandez, Daniel HAZIZA, Francisco Massa, Alaaeldin El-Nouby, Mido Assran, Nicolas Ballas, Wojciech Galuba, Russell Howes, Po-Yao Huang, Shang-Wen Li, Ishan Misra, Michael Rabbat, Vasu Sharma, Gabriel Synnaeve, Hu~Xu, Herve Jegou, Julien Mairal, Patrick Labatut, Armand Joulin, and Piotr Bojanowski.
\newblock {DINO}v2: Learning robust visual features without supervision.
\newblock \emph{Transactions on Machine Learning Research}, 2024.
\newblock ISSN 2835-8856.
\newblock URL \url{https://openreview.net/forum?id=a68SUt6zFt}.
\newblock Featured Certification.

\bibitem[Radford et~al.(2021)Radford, Kim, Hallacy, Ramesh, Goh, Agarwal, Sastry, Askell, Mishkin, Clark, et~al.]{radford2021learning}
Alec Radford, Jong~Wook Kim, Chris Hallacy, Aditya Ramesh, Gabriel Goh, Sandhini Agarwal, Girish Sastry, Amanda Askell, Pamela Mishkin, Jack Clark, et~al.
\newblock Learning transferable visual models from natural language supervision.
\newblock In \emph{ICML}, 2021.

\bibitem[Roth et~al.(2022)Roth, Pemula, Zepeda, Schölkopf, Brox, and Gehler]{PatchCore}
Karsten Roth, Latha Pemula, Joaquin Zepeda, Bernhard Schölkopf, Thomas Brox, and Peter Gehler.
\newblock Towards total recall in industrial anomaly detection.
\newblock In \emph{CVPR}, 2022.

\bibitem[Rudolph et~al.(2021)Rudolph, Wandt, and Rosenhahn]{differnet}
Marco Rudolph, Bastian Wandt, and Bodo Rosenhahn.
\newblock Same same but differnet: Semi-supervised defect detection with normalizing flows.
\newblock In \emph{WACV}, 2021.

\bibitem[Sim{\'e}oni et~al.(2025)Sim{\'e}oni, Vo, Seitzer, Baldassarre, Oquab, Jose, Khalidov, Szafraniec, Yi, Ramamonjisoa, et~al.]{simeoni2025dinov3}
Oriane Sim{\'e}oni, Huy~V Vo, Maximilian Seitzer, Federico Baldassarre, Maxime Oquab, Cijo Jose, Vasil Khalidov, Marc Szafraniec, Seungeun Yi, Micha{\"e}l Ramamonjisoa, et~al.
\newblock Dinov3, 2025.
\newblock URL \url{https://arxiv.org/abs/2508.10104}.

\bibitem[Str{\"a}ter et~al.(2024)Str{\"a}ter, Salehi, Gavves, Snoek, and Asano]{strater2024generalad}
Luc~PJ Str{\"a}ter, Mohammadreza Salehi, Efstratios Gavves, Cees~GM Snoek, and Yuki~M Asano.
\newblock Generalad: Anomaly detection across domains by attending to distorted features.
\newblock In \emph{ECCV}, 2024.

\bibitem[Tien et~al.(2023)Tien, Nguyen, Tran, Huy, Duong, Nguyen, and Truong]{RD++}
Tran~Dinh Tien, Anh~Tuan Nguyen, Nguyen~Hoang Tran, Ta~Duc Huy, Soan Duong, Chanh D~Tr Nguyen, and Steven~QH Truong.
\newblock Revisiting reverse distillation for anomaly detection.
\newblock In \emph{CVPR}, 2023.

\bibitem[Wang et~al.(2025)Wang, Chen, Karaev, Vedaldi, Rupprecht, and Novotny]{wang2025vggt}
Jianyuan Wang, Minghao Chen, Nikita Karaev, Andrea Vedaldi, Christian Rupprecht, and David Novotny.
\newblock Vggt: Visual geometry grounded transformer.
\newblock In \emph{CVPR}, 2025.

\bibitem[Yan et~al.(2024)Yan, Fang, Lv, and Su]{yan2024anomalysd}
Zhenyu Yan, Qingqing Fang, Wenxi Lv, and Qinliang Su.
\newblock Anomalysd: Few-shot multi-class anomaly detection with stable diffusion model, 2024.
\newblock URL \url{https://arxiv.org/abs/2408.01960}.

\bibitem[Yang et~al.(2023)Yang, Wu, and Feng]{yang2023memseg}
Minghui Yang, Peng Wu, and Hui Feng.
\newblock Memseg: A semi-supervised method for image surface defect detection using differences and commonalities.
\newblock \emph{Engineering Applications of Artificial Intelligence}, 119:\penalty0 105835, 2023.

\bibitem[Yi \& Yoon(2020)Yi and Yoon]{patchsvdd}
Jihun Yi and Sungroh Yoon.
\newblock Patch svdd: Patch-level svdd for anomaly detection and segmentation.
\newblock In \emph{ACCV}, 2020.

\bibitem[You et~al.(2022)You, Cui, Shen, Yang, Lu, Zheng, and Le]{you2022unified}
Zhiyuan You, Lei Cui, Yujun Shen, Kai Yang, Xin Lu, Yu~Zheng, and Xinyi Le.
\newblock A unified model for multi-class anomaly detection.
\newblock In \emph{NeurIPS}, 2022.

\bibitem[Zagoruyko \& Komodakis(2016)Zagoruyko and Komodakis]{zagoruyko2016wide}
Sergey Zagoruyko and Nikos Komodakis.
\newblock Wide residual networks.
\newblock In \emph{BMVC}, 2016.

\bibitem[Zavrtanik et~al.(2021)Zavrtanik, Kristan, and Sko{\v{c}}aj]{zavrtanik2021draem}
Vitjan Zavrtanik, Matej Kristan, and Danijel Sko{\v{c}}aj.
\newblock Draem-a discriminatively trained reconstruction embedding for surface anomaly detection.
\newblock In \emph{ICCV}, 2021.

\bibitem[Zavrtanik et~al.(2022)Zavrtanik, Kristan, and Sko{\v{c}}aj]{zavrtanik2022dsr}
Vitjan Zavrtanik, Matej Kristan, and Danijel Sko{\v{c}}aj.
\newblock Dsr--a dual subspace re-projection network for surface anomaly detection.
\newblock In \emph{ECCV}, 2022.

\bibitem[Zhai et~al.(2023{\natexlab{a}})Zhai, {\"O}rnek, Wu, Di, Tombari, Navab, and Busam]{zhai2023commonscenes}
Guangyao Zhai, Evin~P{\i}nar {\"O}rnek, Shun-Cheng Wu, Yan Di, Federico Tombari, Nassir Navab, and Benjamin Busam.
\newblock Commonscenes: Generating commonsense 3d indoor scenes with scene graph diffusion.
\newblock In \emph{NeurIPS}, 2023{\natexlab{a}}.

\bibitem[Zhai et~al.(2024)Zhai, {\"O}rnek, Chen, Liao, Di, Navab, Tombari, and Busam]{zhai2024echoscene}
Guangyao Zhai, Evin~P{\i}nar {\"O}rnek, Dave~Zhenyu Chen, Ruotong Liao, Yan Di, Nassir Navab, Federico Tombari, and Benjamin Busam.
\newblock Echoscene: Indoor scene generation via information echo over scene graph diffusion.
\newblock In \emph{ECCV}, 2024.

\bibitem[Zhai et~al.(2023{\natexlab{b}})Zhai, Mustafa, Kolesnikov, and Beyer]{zhai2023sigmoid}
Xiaohua Zhai, Basil Mustafa, Alexander Kolesnikov, and Lucas Beyer.
\newblock Sigmoid loss for language image pre-training.
\newblock In \emph{ICCV}, 2023{\natexlab{b}}.

\bibitem[Zhang et~al.(2025{\natexlab{a}})Zhang, Wang, Zeng, Wu, and Jiang]{zhang2023diffusionad}
Hui Zhang, Zheng Wang, Dan Zeng, Zuxuan Wu, and Yu-Gang Jiang.
\newblock Diffusionad: Norm-guided one-step denoising diffusion for anomaly detection.
\newblock \emph{IEEE Transactions on Pattern Analysis and Machine Intelligence}, 2025{\natexlab{a}}.

\bibitem[Zhang et~al.(2025{\natexlab{b}})Zhang, Wang, Jin, and Huang]{zhang2025towards}
Jinjin Zhang, Guodong Wang, Yizhou Jin, and Di~Huang.
\newblock Towards training-free anomaly detection with vision and language foundation models.
\newblock In \emph{CVPR}, 2025{\natexlab{b}}.

\bibitem[Zhang et~al.(2024)Zhang, Xu, and Zhou]{zhang2024realnet}
Ximiao Zhang, Min Xu, and Xiuzhuang Zhou.
\newblock Realnet: A feature selection network with realistic synthetic anomaly for anomaly detection.
\newblock In \emph{CVPR}, 2024.

\bibitem[Zhang et~al.(2023)Zhang, Li, Li, Huang, Shan, and Chen]{destseg}
Xuan Zhang, Shiyu Li, Xi~Li, Ping Huang, Jiulong Shan, and Ting Chen.
\newblock Destseg: Segmentation guided denoising student-teacher for anomaly detection.
\newblock In \emph{CVPR}, 2023.

\bibitem[Zhao(2023)]{zhao2023omnial}
Ying Zhao.
\newblock Omnial: A unified cnn framework for unsupervised anomaly localization.
\newblock In \emph{CVPR}, 2023.

\bibitem[Zhou et~al.(2022)Zhou, Loy, and Dai]{zhou2022extract}
Chong Zhou, Chen~Change Loy, and Bo~Dai.
\newblock Extract free dense labels from clip.
\newblock In \emph{ECCV}, 2022.

\bibitem[Zhou et~al.(2025)Zhou, Pan, LeCun, and Pinto]{zhou2024dino}
Gaoyue Zhou, Hengkai Pan, Yann LeCun, and Lerrel Pinto.
\newblock Dino-wm: World models on pre-trained visual features enable zero-shot planning.
\newblock In \emph{ICML}, 2025.

\bibitem[Zhou et~al.(2024)Zhou, Pang, Tian, He, and Chen]{zhouanomalyclip}
Qihang Zhou, Guansong Pang, Yu~Tian, Shibo He, and Jiming Chen.
\newblock Anomalyclip: Object-agnostic prompt learning for zero-shot anomaly detection.
\newblock In \emph{ICLR}, 2024.

\bibitem[Zhu \& Pang(2024)Zhu and Pang]{zhu2024generalistanomalydetectionincontext}
Jiawen Zhu and Guansong Pang.
\newblock Toward generalist anomaly detection via in-context residual learning with few-shot sample prompts.
\newblock In \emph{CVPR}, 2024.

\bibitem[Zou et~al.(2022)Zou, Jeong, Pemula, Zhang, and Dabeer]{visa}
Yang Zou, Jongheon Jeong, Latha Pemula, Dongqing Zhang, and Onkar Dabeer.
\newblock Spot-the-difference self-supervised pre-training for anomaly detection and segmentation.
\newblock In \emph{ECCV}, 2022.

\end{thebibliography}
}

\appendix
% WARNING: do not forget to delete the supplementary pages from your submission 
\clearpage
{\large
\textbf{Supplementary Material}
}
% \setcounter{page}{1}
% \maketitlesupplementary

\smallskip
In this material, we provide more details about the ablation of foundation encoders and additional quantitative and qualitative results.

\section{Experimental Detail for Encoder Ablation}
\label{encoder_ablation_detail}
We performed experiments with various encoders under a 1-shot setting on MVTec-AD. To maintain consistency with the default encoder DINOv3 of \name{}, use the same sampled training instances across different encoders. The projector was uniformly maintained at a depth of 6, and the learning rate was consistently set to 0.001 for all cases. 

To ensure a fair comparison, we use ViT-Base for DINOv3, DINOv2, DINO, SigLIP, and CLIP, all in the third-to-last layer. We conducted the experiment for DINOSigLIP using ViT-L due to availability. However, since different encoders require varying default image input sizes and patch sizes, the input patch numbers for the projector was adjusted accordingly: (1) For DINOv2 and DINO ViT-B, the input images were resized to 518, with a patch size of 14. (2) For DINOv3 and SigLIP ViT-B, the input images were resized to 512, with a patch size of 16. (3) For CLIP ViT-B, the default image size was 224 with a patch size of 16. (4) For DINOSigLIP, we employed the implementation from Prismatic VLMs~\citep{karamcheti2024prismatic} \footnote{\url{https://github.com/tri-ml/prismatic-vlms}}, which combines DINOv2 ViT-L and SigLIP ViT-SO embeddings with a linear projection at an input resolution of 384. (5) For WideResNet, we adopted the methodology identical to PatchCore~\citep{PatchCore}.

% \begin{figure*}[h]
%     \centering
%     \includegraphics[width=0.9\linewidth]{figures/failure.png}
%     \caption{Typical failure cases of \name{}.}
%     \label{fig:failure_case}
% \end{figure*}

\section{Additional Results}
\label{additional}

\begin{table}[h]
\captionsetup{justification=centering,singlelinecheck=false}
\centering
\caption{ \small Results on MVTec-AD and VisA under multiple seeds.}
    \centering
    \footnotesize
    \resizebox{0.85\textwidth}{!}{
    \begin{tabular}{cccccccccc}
    \toprule
    \multirow{2}{*}{\textbf{Setting}} & \multirow{2}{*}{\textbf{Seed}} 
      & \multicolumn{4}{c}{\textbf{MVTec-AD}} 
      & \multicolumn{4}{c}{\textbf{VisA}} \\
    \cmidrule(l{1em}r{1em}){3-6}
    \cmidrule(l{1em}r{1em}){7-10}
     &  & {I-AUROC} & {AUPR} & {P-AUROC} & {PRO}
        & {I-AUROC} & {AUPR} & {P-AUROC} & {PRO} \\
    \midrule

    %--------- 1-shot + Multi-class ---------
    \multirow{4}{*}{\shortstack{1-shot}}
    & 1     & 96.0 & 97.8 & 96.9 & 93.1 & 92.5 & 91.5 & 99.7 & 97.8 \\
    & 2     & 96.0 & 97.6 & 96.9 & 92.8 & 91.9 & 91.9 & 99.7 & 98.2 \\
    & 3     & 96.4 & 98.2 & 96.5 & 92.6 & 93.4 & 92.7 & 99.7 & 98.1 \\
    \cmidrule(l{1em}r{0em}){2-10}
    & \best{Avg}     & \best96.1 & \best97.9 & \best96.8 & \best92.8 & \best92.6 & \best92.0 & \best99.7 & \best98.0 \\

    \midrule
    
    \multirow{4}{*}{\shortstack{2-shot}}
    & 1     & {96.8} & {98.3} & {97.0} & {93.3} & 93.9 & 93.2 & 99.7 & 98.1 \\
    & 2     & 96.9 & 98.2 & 97.2 & 93.7 & 93.5 & 93.3 & 99.7 & 98.3 \\
    & 3     & 96.6 & 98.3 & 96.9 & 92.9 & 94.0 & 93.3 & 99.7 & 98.2 \\
    \cmidrule(l{1em}r{0em}){2-10}
    & \best{Avg}     & \best96.8 & \best98.3 & \best97.0 & \best93.3 & \best93.8 & \best93.3 & \best99.7 & \best98.2 \\

    \midrule
    
    \multirow{4}{*}{\shortstack{4-shot}}
    & 1     & {97.2} & {98.6} & {97.2} & {93.5} & 94.2 & 93.4 & 99.7 & 98.2 \\
    & 2     & 97.2 & 98.6 & 97.3 & 93.7 & 94.5 & 94.6 & 99.7 & 98.4 \\
    & 3     & 97.0 & 98.5 & 97.1 & 93.4 & 94.6 & 93.9 & 99.7 & 98.6 \\
    \cmidrule(l{1em}r{0em}){2-10}
    & \best{Avg}     & \best97.1 & \best98.6 & \best97.2 & \best93.5 & \best94.4 & \best94.0 & \best99.7 & \best98.4 \\
    \bottomrule
    \end{tabular}}
    \label{table:ablation_multiple_run}
\end{table}

\subsection{Multiple Run Results}
\label{multiple_run}
To demonstrate the robustness and reliability of \name{}, we repeated the experiments in three random seeds on the MVTec-AD dataset. The detailed experimental results are summarized in \autoref{table:ablation_multiple_run}. 
%The first run corresponds to the originally reported result, achieving an I-AUROC of 94.5\%, AUPR of 97.2\%, P-AUROC of 96.7\%, and PRO of 92.3\%. In four additional independent runs, we observed consistent performance with minor fluctuations across the metrics: I-AUROC ranged from 93.0\% to 94.3\%, AUPR varied between 96.4\% and 97.1\%, P-AUROC exhibited high stability between 96.4\% and 96.8\%, and PRO scores slightly fluctuated within 91.7\% to 92.0\%. 
These results indicate that \name{} consistently maintains strong generalization capability and stable anomaly detection performance even under limited data conditions.

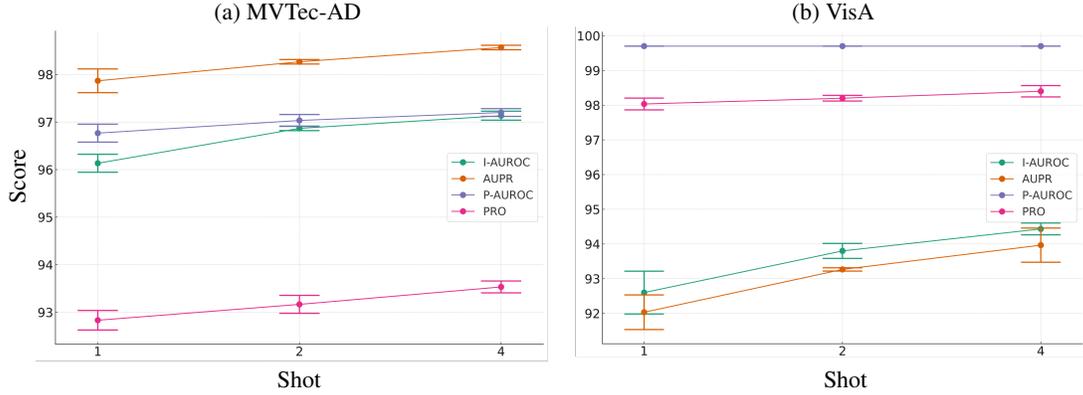
\begin{figure}[h]
  \centering
  \vspace{14px}
  \begin{overpic}[width=\linewidth,trim=0cm 0 0 0cm,clip]{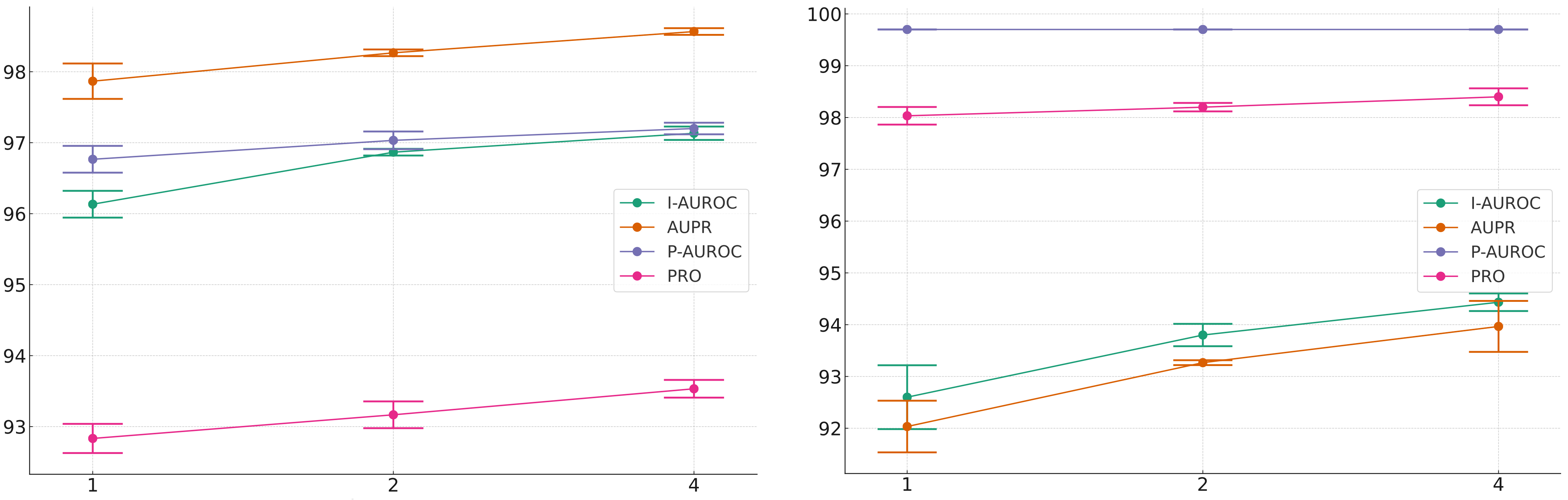}
    
    \put(17,32.5){\color{black}{\small (a) MVTec-AD}}
    \put(72,32.5){\color{black}{\small (b) VisA}}

    \put(-2.5,15){\rotatebox{90}{\small Score}}

    \put(23,-2.5){\color{black}{\small Shot}}
    \put(75,-2.5){\color{black}{\small Shot}}

  \end{overpic}
  \vspace{0.5px}
    \caption{\small Mean and standard deviation of four evaluation metrics across 1-, 2-, and 4-shot settings on MVTec-AD and VisA.
    }
    \label{fig:error_bar}
    \vspace{-3mm}
\end{figure}

All four metrics consistently improve as the number of shots increases, particularly on MVTec-AD, indicating that our method can effectively leverage additional supervision to enhance anomaly detection performance.
As shown in~\autoref{fig:error_bar}, the standard deviations are generally small across runs, demonstrating strong stability and robustness. On VisA, interestingly, the P-AUROC remains nearly saturated at $99.7$ across all settings, showing that the method performs highly reliably at the pixel level.
I-AUROC and AUPR exhibit steeper gains on MVTec-AD, likely due to its more diverse visual patterns, where additional examples provide more substantial benefit, whereas VisA benefits from more homogeneous object categories.
Although the PRO metric shows slight fluctuations, it consistently stays high, underscoring the method’s effectiveness in localizing anomalies with pixel-level precision.
Overall, the results demonstrate our method’s robustness, consistency, and scalability in few-shot settings, making it well-suited for real-world industrial scenarios with limited annotations.

\subsection{Class-Wise Quantitative Results}
\label{more_quantitative_Results}
We report the class-wise quantitative results of \name{} with DINOv3 under 1, 2, and 4 shot settings on MVTec-AD and VisA in detail in \autoref{table: Mvtec_all_results} and \autoref{table: VisA_all_results}.
\vspace{-1mm}

\begin{table*}[h]
    \centering
    \caption{\small Results on MVTec-AD for 1-shot, 2-shot, and 4-shot. Metrics include I-AUROC (\%), AUPR (\%) and pixel-level P-AUROC (\%), PRO (\%).}
    \label{table: Mvtec_all_results}
    \resizebox{0.95\textwidth}{!}{%
    \begin{tabular}{lcccccccccccc}
    \toprule
    \multirow{2}{*}{\textbf{Class}} 
    & \multicolumn{4}{c}{\textbf{1-shot}} 
    & \multicolumn{4}{c}{\textbf{2-shot}} 
    & \multicolumn{4}{c}{\textbf{4-shot}} 
    \\
    \cmidrule(lr){2-5}\cmidrule(lr){6-9}\cmidrule(lr){10-13}
        & \textbf{I-AUROC} & \textbf{AUPR} & \textbf{P-AUROC} & \textbf{PRO}
        & \textbf{I-AUROC} & \textbf{AUPR} & \textbf{P-AUROC} & \textbf{PRO}
        & \textbf{I-AUROC} & \textbf{AUPR} & \textbf{P-AUROC} & \textbf{PRO} \\
    \midrule
    Bottle      & 100.0 & 100.0 & 98.6 & 96.2 & 100.0 & 100.0 & 98.6 & 95.8 & 100.0 & 100.0 & 98.6 & 95.9 \\
    Cable       & 89.6  & 94.1  & 94.7 & 89.2 & 92.3  & 96.3  & 95.0 & 89.9 & 92.4  & 96.4  & 94.7 & 89.6 \\
    Capsule     & 89.5  & 97.5  & 98.6 & 94.3 & 89.6  & 97.6  & 98.8 & 94.9 & 90.1  & 97.7  & 98.8 & 94.8 \\
    Carpet      & 100.0 & 100.0 & 99.5 & 98.3 & 100.0 & 100.0 & 99.4 & 98.3 & 100.0 & 100.0 & 99.5 & 98.2 \\
    Grid        & 99.8  & 99.9  & 99.2 & 96.1 & 99.9  & 99.9  & 99.3 & 96.6 & 100.0 & 100.0 & 99.3 & 96.8 \\
    Hazelnut    & 99.3  & 99.7  & 99.5 & 93.0 & 99.9  & 99.9  & 99.5 & 93.2 & 99.8  & 99.8  & 99.5 & 93.4 \\
    Leather     & 100.0 & 100.0 & 99.4 & 98.7 & 100.0 & 100.0 & 99.4 & 98.7 & 100.0 & 100.0 & 99.4 & 98.4 \\
    Metal\_nut  & 99.8  & 100.0 & 92.5 & 91.8 & 99.9  & 100.0 & 94.3 & 93.3 & 100.0 & 100.0 & 96.3 & 95.3 \\
    Pill        & 97.7  & 99.6  & 94.5 & 97.0 & 97.8  & 99.6  & 94.6 & 97.1 & 98.1  & 99.7  & 95.0 & 96.7 \\
    Screw       & 84.8  & 94.3  & 98.4 & 92.6 & 87.6  & 95.2  & 98.7 & 92.8 & 89.0  & 95.7  & 98.7 & 93.3 \\
    Tile        & 100.0 & 100.0 & 97.2 & 95.4 & 100.0 & 100.0 & 97.3 & 95.1 & 100.0 & 100.0 & 97.4 & 94.9 \\
    Toothbrush  & 99.8  & 99.9  & 99.2 & 94.1 & 99.2  & 99.7  & 99.2 & 95.2 & 100.0 & 100.0 & 99.4 & 95.6 \\
    Transistor  & 86.3  & 85.0  & 85.0 & 63.3 & 89.2  & 86.6  & 86.6 & 63.8 & 92.5  & 90.0  & 87.4 & 65.7 \\
    Wood        & 95.4  & 97.9  & 96.2 & 96.0 & 96.9  & 98.7  & 96.2 & 95.9 & 95.2  & 98.6  & 96.2 & 96.1 \\
    Zipper      & 99.8  & 99.9  & 99.0 & 96.8 & 99.8  & 99.9  & 99.0 & 97.1 & 99.9  & 100.0 & 99.1 & 97.0 \\
    \midrule
    \textbf{Average}
                & 96.1  & 97.9  & 96.8 & 92.8 & 96.8  & 98.3  & 97.0 & 93.3 & 97.1  & 98.6  & 97.2 & 93.5 \\
    \bottomrule
    \end{tabular}%
    }
    \vspace{-2mm}
\end{table*}

\begin{table*}[h]
    \centering
    \caption{\small Results of \name{} on VisA for 1-shot, 2-shot, and 4-shot. Metrics include I-AUROC (\%), AUPR (\%) and pixel-level P-AUROC (\%), PRO (\%).}
    \label{table: VisA_all_results}
    \resizebox{0.95\textwidth}{!}{%
    \begin{tabular}{lcccccccccccc}
    \toprule
    \multirow{2}{*}{\textbf{Class}} 
    & \multicolumn{4}{c}{\textbf{1-shot}} 
    & \multicolumn{4}{c}{\textbf{2-shot}} 
    & \multicolumn{4}{c}{\textbf{4-shot}} 
    \\
    \cmidrule(lr){2-5}\cmidrule(lr){6-9}\cmidrule(lr){10-13}
       & \textbf{I-AUROC} & \textbf{AUPR} & \textbf{P-AUROC} & \textbf{PRO}
       & \textbf{I-AUROC} & \textbf{AUPR} & \textbf{P-AUROC} & \textbf{PRO}
       & \textbf{I-AUROC} & \textbf{AUPR} & \textbf{P-AUROC} & \textbf{PRO} \\
    \midrule
    Candle      & 94.4  & 93.2  & 99.5  & 98.4  & 94.7  & 93.3  & 99.6  & 98.7  & 94.3  & 94.1  & 99.5  & 98.4  \\
    Capsules    & 98.9  & 99.3  & 99.8  & 99.0  & 99.0  & 99.4  & 99.8  & 99.0  & 99.1  & 99.4  & 99.8  & 99.0  \\
    Cashew      & 97.4  & 98.6  & 99.6  & 98.8  & 97.0  & 98.5  & 99.6  & 98.5  & 97.6  & 98.7  & 99.6  & 98.5  \\
    Chewinggum  & 98.0  & 99.1  & 99.9  & 99.0  & 98.5  & 99.3  & 99.9  & 99.1  & 98.5  & 99.3  & 99.8  & 98.9  \\
    Fryum       & 96.8  & 98.5  & 99.8  & 98.0  & 96.8  & 98.5  & 99.8  & 98.2  & 97.2  & 98.6  & 99.7  & 98.1  \\
    Macaroni1   & 93.1  & 92.2  & 99.5  & 98.0  & 93.8  & 92.9  & 99.6  & 98.3  & 94.0  & 93.0  & 99.6  & 98.3  \\
    Macaroni2   & 72.3  & 64.9  & 99.9  & 99.8  & 75.2  & 68.0  & 99.9  & 99.8  & 77.2  & 71.1  & 99.9  & 99.9  \\
    Pcb1        & 94.1  & 90.8  & 99.5  & 98.1  & 95.3  & 92.9  & 99.5  & 98.2  & 96.9  & 94.6  & 99.6  & 98.5  \\
    Pcb2        & 92.6  & 90.5  & 99.6  & 97.6  & 93.9  & 92.1  & 99.6  & 97.6  & 93.9  & 91.8  & 99.7  & 98.0  \\
    Pcb3        & 82.0  & 84.7  & 99.8  & 99.1  & 87.8  & 90.0  & 99.9  & 99.4  & 90.5  & 91.6  & 99.9  & 99.5  \\
    Pcb4        & 96.6  & 94.8  & 99.7  & 93.2  & 97.2  & 96.6  & 99.7  & 94.7  & 97.5  & 96.5  & 99.7  & 95.2  \\
    Pipe\_fryum & 95.0  & 97.5  & 99.7  & 96.9  & 96.1  & 98.0  & 99.7  & 97.0  & 96.9  & 98.5  & 99.7  & 96.8  \\
    \midrule
    \textbf{Average}  
                & 92.6  & 92.0  & 99.7  & 98.0  
                & 93.8  & 93.3  & 99.7  & 98.2  
                & 94.4  & 94.0  & 99.7  & 98.4  \\
    \bottomrule
    \end{tabular}%
    }
\end{table*}

\subsection{Additional Qualitative Comparisons}
\label{add:logsad}
\begin{figure}[h]
  \centering
  \vspace{14px}
  \begin{overpic}[width=0.7\linewidth,trim=0cm 0 0 0cm,clip]{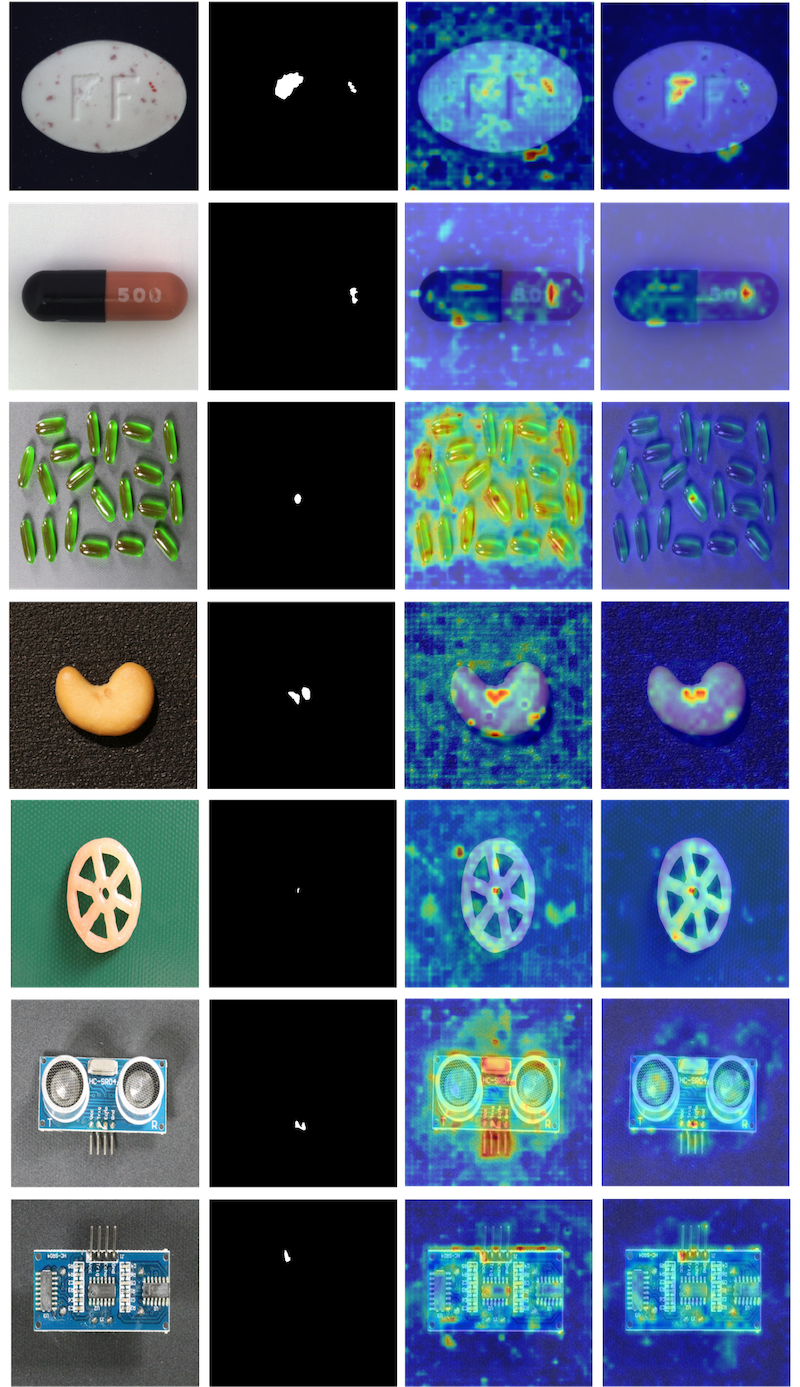}
    
    \put(4,101.5){\color{black}\small {Abnormal}}
    \put(17.1,101.5){\color{black}\small {Ground Truth}}
    \put(33,101.5){\color{black}\small {LogSAD}}
    \put(46.3,101.5){\color{black}\small {\name{}}}
    
    \put(-2.5,92){\rotatebox{90}{\small Pill}}
    \put(-2.5,76.2){\rotatebox{90}{\small Capsule}}
    \put(-2.5,61.5){\rotatebox{90}{\small Capsules}}
    \put(-2.5,47.3){\rotatebox{90}{\small Cashew}}
    \put(-2.5,33){\rotatebox{90}{\small Fryum}}
    \put(-2.5,19.2){\rotatebox{90}{\small Pcb1}}
    \put(-2.5,5.3){\rotatebox{90}{\small Pcb2}}

  \end{overpic}
  \vspace{0.5px}
    \caption{\small Visual comparisons with LogSAD~\citep{zhang2025towards}.
    }
    \label{fig:logsad}
    \vspace{-3mm}
\end{figure}

We provide several visual comparisons against LogSAD~\citep{zhang2025towards} in~\autoref{fig:logsad}. It clearly shows that \name{} localizes anomaly precisely, while producing significantly less noise.

\subsection{Class-Wise Comparison to Other Methods}
The class-wise comparisons are presented with I-AUROC  in \autoref{table: Mvtec_I_AUROC_results}-\autoref{table: VisA_I_AUROC_results}, AUPR in \autoref{table: Mvtec_AUPR_results}-\autoref{table: VisA_AUPR_results}, P-AUROC in \autoref{table: Mvtec_P_AUROC_results}-\autoref{table: VisA_P_AUROC_results}, and PRO in \autoref{table: Mvtec_PRO_results}-\autoref{table: VisA_PRO_results}.

\begin{table*}[h]
 \vspace{-2mm}
    \centering
    \caption{\small I-AUROC (\%) results of MVTec-AD for 1-shot, 2-shot, and 4-shot.}
    \vspace{-2mm}
    \label{table: Mvtec_I_AUROC_results}
    \resizebox{0.95\textwidth}{!}{%
    \begin{tabular}{lcccccccccccc}
    \toprule
    \multirow{2}{*}{\textbf{Class}} 
    & \multicolumn{4}{c}{\textbf{1-shot}} 
    & \multicolumn{4}{c}{\textbf{2-shot}} 
    & \multicolumn{4}{c}{\textbf{4-shot}} 
    \\
    \cmidrule(lr){2-5}\cmidrule(lr){6-9}\cmidrule(lr){10-13}
        & \textbf{WinCLIP} & \textbf{PromptAD} & \textbf{IIPAD} & \textbf{Ours}  
        & \textbf{WinCLIP} & \textbf{PromptAD} & \textbf{IIPAD} & \textbf{Ours} 
        & \textbf{WinCLIP} & \textbf{PromptAD} & \textbf{IIPAD} & \textbf{Ours} \\
    \midrule
    Bottle      
        & 98.9 & 98.6 & 99.7  & 100.0
        & 99.2 & 100.0 & 99.8  & 100.0
        & 99.2 & 99.0 & 99.1  & 99.1 \\
    Cable       
        & 78.0 & 83.6 & 92.8  & 89.6
        & 83.9 & 87.2 & 92.1  & 92.3
        & 82.3 & 88.7 & 95.4  & 92.4\\
    Capsule     
        & 75.5 & 64.2 & 80.5  & 89.5
        & 65.5 & 65.3 & 91.8  & 89.6
        & 80.1 & 93.4 & 94.5  & 90.1 \\
    Carpet      
        & 99.9 & 100.0 & 100.0  & 100.0
        & 99.9 & 100.0 & 100.0  & 100.0
        & 99.9 & 100.0 & 100.0  & 100.0\\
    Grid        
        & 99.6 & 98.8 & 97.0  & 99.8
        & 99.2 & 97.4 & 97.0  & 99.9
        & 99.5 & 100.0 & 96.0  & 100.0\\
    Hazelnut    
        & 94.9 & 98.4 & 98.0  & 99.3
        & 95.2 & 99.8 & 98.5  & 99.9
        & 94.7 & 99.0 & 98.5  & 99.8 \\
    Leather     
        & 100.0 & 100.0 & 100.0  & 100.0
        & 100.0 & 100.0 & 100.0  & 100.0
        & 100.0 & 100.0 & 100.0  & 100.0 \\
    Metal\_nut   
        & 98.0 & 97.6 & 99.4  & 99.8
        & 97.8 & 96.2 & 99.7  & 99.9
        & 98.9 & 100.0 & 99.9  & 100.0 \\
    Pill        
        & 88.9 & 87.9 & 96.6  & 97.7
        & 91.8 & 89.1 & 96.0  & 97.8
        & 91.1 & 90.4 & 96.6  & 98.1 \\
    Screw       
        & 85.1 & 74.0 & 76.8  & 84.8
        & 82.7 & 81.2 & 81.5  & 87.6
        & 84.4 & 84.2 & 82.1  & 89.0 \\
    Tile        
        & 100.0 & 99.8 & 99.7  & 100.0
        & 100.0 & 99.3 & 99.5  & 100.0
        & 100.0 & 99.2 & 99.9  & 100.0 \\
    Toothbrush  
        & 94.2 & 94.4 & 91.9  & 99.8
        & 93.9 & 100.0 & 92.5  & 99.2
        & 98.1 & 98.8 & 92.5  & 100.0 \\
    Transistor  
        & 85.5 & 73.7 & 91.4  & 86.3
        & 85.4 & 87.2 & 90.4  & 89.2
        & 85.6 & 94.4 & 91.2  & 92.5\\
    Wood        
        & 98.7 & 98.6 & 99.4  & 95.4
        & 98.9 & 98.9 & 99.2  & 96.9
        & 98.9 & 99.2 & 99.6  & 95.2 \\
    Zipper      
        & 94.9 & 95.3 & 89.4  & 99.8
        & 97.2 & 93.5 & 95.5  & 99.8
        & 97.0 & 95.8 & 96.0  & 99.9 \\
    \midrule
    \textbf{Mean} 
        & 92.8 & 86.3 & 94.2  & 96.1
        & 92.7 & 89.2 & 95.6  & 96.8
        & 94.0 & 90.6 & 96.1  & 97.1 \\
    \bottomrule
    \end{tabular}%
    }
\vspace{-2mm}
\end{table*}

\begin{table*}[h]
    \centering
    \caption{\small I-AUROC (\%) results of VisA for 1-shot, 2-shot, and 4-shot.}
    \vspace{-2mm}
    \label{table: VisA_I_AUROC_results}
    \resizebox{0.95\textwidth}{!}{%
    \begin{tabular}{lcccccccccccc}
    \toprule
    \multirow{2}{*}{\textbf{Class}} 
    & \multicolumn{4}{c}{\textbf{1-shot}} 
    & \multicolumn{4}{c}{\textbf{2-shot}} 
    & \multicolumn{4}{c}{\textbf{4-shot}} 
    \\
    \cmidrule(lr){2-5}\cmidrule(lr){6-9}\cmidrule(lr){10-13}
       & \textbf{WinCLIP} & \textbf{PromptAD} & \textbf{IIPAD} & \textbf{Ours}  
       & \textbf{WinCLIP} & \textbf{PromptAD} & \textbf{IIPAD} & \textbf{Ours}  
       & \textbf{WinCLIP} & \textbf{PromptAD} & \textbf{IIPAD} & \textbf{Ours}  \\
    \midrule
    Candle      
        & 96.3 & 91.8 & 91.9 & 94.4  
        & 96.4 & 92.0 & 95.5 & 94.7  
        & 96.9 & 92.9 & 95.9 & 94.3  \\
    Capsules    
        & 79.3 & 83.2 & 88.9 & 98.9  
        & 81.6 & 78.7 & 90.3 & 99.0  
        & 83.0 & 81.7 & 90.5 & 99.1  \\
    Cashew      
        & 93.9 & 88.9 & 85.6 & 97.4  
        & 92.6 & 89.6 & 86.7 & 97.0  
        & 92.6 & 88.0 & 91.2 & 97.6  \\
    Chewinggum  
        & 97.9 & 97.3 & 97.7 & 98.0  
        & 98.1 & 97.1 & 97.8 & 98.5  
        & 98.4 & 98.1 & 98.0 & 98.5  \\
    Fryum       
        & 92.8 & 88.0 & 89.9 & 96.8  
        & 90.1 & 85.7 & 92.7 & 96.8  
        & 91.6 & 90.6 & 93.3 & 97.2  \\
    Macaroni1   
        & 81.9 & 87.3 & 85.1 & 93.1  
        & 86.4 & 87.4 & 84.7 & 93.8  
        & 86.9 & 89.1 & 88.4 & 94.0  \\
    Macaroni2   
        & 78.1 & 60.8 & 75.5 & 72.3  
        & 76.8 & 74.9 & 76.1 & 75.2  
        & 79.0 & 80.5 & 78.1 & 77.2  \\
    Pcb1        
        & 83.8 & 83.0 & 83.5 & 94.1  
        & 85.5 & 82.9 & 86.5 & 95.3  
        & 86.0 & 86.1 & 85.2 & 96.9  \\
    Pcb2        
        & 58.4 & 77.9 & 72.6 & 92.6  
        & 56.8 & 84.4 & 75.2 & 93.9  
        & 59.4 & 81.1 & 75.5 & 93.9  \\
    Pcb3        
        & 64.9 & 79.9 & 71.8 & 82.0  
        & 67.7 & 71.7 & 70.7 & 87.8  
        & 65.6 & 87.1 & 74.7 & 90.5  \\
    Pcb4        
        & 72.1 & 96.5 & 82.9 & 96.6  
        & 73.6 & 96.0 & 84.4 & 97.2  
        & 70.7 & 85.3 & 88.7 & 97.5  \\
    Pipefryum   
        & 98.2 & 98.9 & 99.8 & 95.0  
        & 98.5 & 99.6 & 99.9 & 96.1  
        & 98.4 & 99.3 & 99.8 & 96.9  \\
    \midrule
    \textbf{Mean} 
        & 83.1 & 80.8 & 85.4 & 92.6  
        & 83.7 & 84.3 & 86.7 & 93.8  
        & 84.1 & 85.7 & 88.3 & 94.4  \\
    \bottomrule
    \end{tabular}%
    }
    \vspace{-2mm}
\end{table*}

\begin{table*}[h]
    \centering
    \caption{\small AUPR (\%) results of MVTec-AD for 1-shot, 2-shot, and 4-shot.}
    \vspace{-2mm}
    \label{table: Mvtec_AUPR_results}
    \resizebox{0.95\textwidth}{!}{%
    \begin{tabular}{lcccccccccccc}
    \toprule
    \multirow{2}{*}{\textbf{Class}} 
    & \multicolumn{4}{c}{\textbf{1-shot}} 
    & \multicolumn{4}{c}{\textbf{2-shot}} 
    & \multicolumn{4}{c}{\textbf{4-shot}} 
    \\
    \cmidrule(lr){2-5}\cmidrule(lr){6-9}\cmidrule(lr){10-13}
        & \textbf{WinCLIP} & \textbf{PromptAD} & \textbf{IIPAD} & \textbf{Ours}  
        & \textbf{WinCLIP} & \textbf{PromptAD} & \textbf{IIPAD} & \textbf{Ours} 
        & \textbf{WinCLIP} & \textbf{PromptAD} & \textbf{IIPAD} & \textbf{Ours} \\
    \midrule
    Bottle      
        & 99.7 & 99.6 & 99.9  & 100.0
        & 99.8 & 100.0 & 100.0  & 100.0
        & 99.8 & 99.7 & 99.7  & 100.0 \\
    Cable       
        & 87.1 & 91.2 & 96.1  & 94.1
        & 90.6 & 92.8 & 95.8  & 96.3
        & 90.0 & 93.6 & 97.5  & 96.4\\
    Capsule     
        & 93.9 & 85.7 & 95.4  & 97.5 
        & 88.3 & 85.5 & 98.3  & 97.6
        & 94.8 & 98.5 & 98.9  & 97.7\\
    Carpet      
        & 100.0 & 100.0 & 100.0  & 100.0
        & 100.0 & 100.0 & 100.0  & 100.0
        & 100.0 & 100.0 & 100.0  & 100.0 \\
    Grid        
        & 99.9 & 99.5 & 99.0  & 99.9
        & 99.7 & 99.1 & 99.0  & 99.9
        & 99.8 & 100.0 & 98.6  & 100.0 \\
    Hazelnut    
        & 97.4 & 99.1 & 99.2  & 99.7
        & 97.5 & 99.9 & 99.4  & 99.9
        & 97.2 & 99.4 & 99.4  & 99.8 \\
    Leather     
        & 100.0 & 100.0 & 100.0 & 100.0 
        & 100.0 & 100.0 & 100.0  & 100.0
        & 100.0 & 100.0 & 100.0  & 100.0 \\
    Metal\_nut   
        & 99.6 & 99.3 & 99.9  & 100.0
        & 99.5 & 98.3 & 99.9  & 100.0
        & 99.8 & 100.0 & 100.0  & 100.0 \\
    Pill        
        & 97.6 & 96.8 & 99.4  & 99.6
        & 98.3 & 97.1 & 99.3  & 99.6
        & 98.2 & 97.7 & 99.4  & 99.7 \\
    Screw       
        & 95.1 & 91.19 & 87.0  & 94.3
        & 93.3 & 93.5 & 92.6  & 95.2
        & 94.0 & 93.7 & 92.0   & 95.7\\
    Tile        
        & 100.0 & 99.9 & 99.9  & 100.0
        & 100.0 & 99.7 & 99.8  & 100.0
        & 100.0 & 99.6 & 99.9  & 100.0 \\
    Toothbrush  
        & 97.7 & 97.7 & 97.3  & 99.9
        & 97.6 & 100.0 & 97.5  & 99.7
        & 99.3 & 99.5 & 97.5  & 100.0 \\
    Transistor  
        & 80.8 & 62.2 & 89.8  & 85.0
        & 81.0 & 77.2 & 88.3  & 86.6
        & 82.6 & 92.2 & 89.2  & 90.0 \\
    Wood        
        & 99.6 & 99.5 & 99.8  & 97.9
        & 99.7 & 99.6 & 99.8  & 98.7
        & 99.7 & 99.7 & 99.9  & 98.6 \\
    Zipper      
        & 98.6 & 98.8 & 95.6  & 99.9
        & 99.2 & 98.2 & 98.7  & 99.9
        & 99.2 & 98.9 & 98.8  & 100.0 \\
    \midrule
    \textbf{Mean} 
        & 96.5 & 93.4 & 97.2  & 97.9 
        & 96.3 & 94.8 & 97.9  & 98.3
        & 97.0 & 96.5 & 98.1  & 98.6\\
    \bottomrule
    \end{tabular}%
    }
    \vspace{-2mm}
\end{table*}

\begin{table*}[h]
    \centering
    \caption{\small AUPR (\%) results of VisA for 1-shot, 2-shot, and 4-shot.}
    \vspace{-2mm}
    \label{table: VisA_AUPR_results}
    \resizebox{0.95\textwidth}{!}{%
    \begin{tabular}{lcccccccccccc}
    \toprule
    \multirow{2}{*}{\textbf{Class}} 
    & \multicolumn{4}{c}{\textbf{1-shot}} 
    & \multicolumn{4}{c}{\textbf{2-shot}} 
    & \multicolumn{4}{c}{\textbf{4-shot}} 
    \\
    \cmidrule(lr){2-5}\cmidrule(lr){6-9}\cmidrule(lr){10-13}
       & \textbf{WinCLIP} & \textbf{PromptAD} & \textbf{IIPAD} & \textbf{Ours}  
       & \textbf{WinCLIP} & \textbf{PromptAD} & \textbf{IIPAD} & \textbf{Ours}  
       & \textbf{WinCLIP} & \textbf{PromptAD} & \textbf{IIPAD} & \textbf{Ours}  \\
    \midrule
    Candle      
        & 96.7 & 90.7 & 94.1 & 93.2  
        & 96.9 & 91.7 & 95.3 & 93.3  
        & 97.3 & 92.8 & 95.5 & 94.1  \\
    Capsules    
        & 87.0 & 90.0 & 94.7 & 99.3  
        & 89.1 & 86.7 & 94.9 & 99.4  
        & 90.0 & 89.0 & 96.0 & 99.4  \\
    Cashew      
        & 97.4 & 95.0 & 95.7 & 98.6  
        & 96.7 & 95.1 & 95.2 & 98.5  
        & 96.8 & 94.7 & 96.0 & 98.7  \\
    Chewinggum  
        & 99.1 & 98.9 & 99.1 & 99.1  
        & 99.2 & 98.8 & 99.2 & 99.3  
        & 99.3 & 99.2 & 99.3 & 99.3  \\
    Fryum       
        & 96.9 & 94.6 & 96.3 & 98.5  
        & 95.4 & 93.9 & 96.8 & 98.5  
        & 96.1 & 95.9 & 97.3 & 98.6  \\
    Macaroni1   
        & 82.8 & 89.7 & 89.3 & 92.2  
        & 87.2 & 89.0 & 90.7 & 92.9  
        & 87.4 & 91.1 & 91.2 & 93.0  \\
    Macaroni2   
        & 80.1 & 61.5 & 79.3 & 64.9  
        & 79.0 & 78.2 & 80.2 & 68.0  
        & 81.9 & 81.2 & 80.6 & 71.1  \\
    Pcb1        
        & 83.7 & 77.3 & 77.4 & 90.8  
        & 84.1 & 79.0 & 78.5 & 92.9  
        & 85.6 & 81.2 & 78.6 & 94.6  \\
    Pcb2        
        & 58.6 & 79.3 & 70.2 & 90.5  
        & 54.6 & 85.5 & 73.6 & 92.1  
        & 61.3 & 80.8 & 73.9 & 91.8  \\
    Pcb3        
        & 66.2 & 81.6 & 73.2 & 84.7  
        & 67.3 & 73.5 & 74.3 & 90.0  
        & 64.6 & 88.0 & 74.6 & 91.6  \\
    Pcb4        
        & 73.8 & 96.1 & 80.9 & 94.8  
        & 70.2 & 94.8 & 81.7 & 96.6  
        & 73.5 & 83.6 & 84.5 & 96.5  \\
    Pipefryum   
        & 99.2 & 99.6 & 99.9 & 97.5  
        & 99.3 & 99.7 & 100  & 98.0  
        & 99.3 & 99.7 & 99.9 & 98.5  \\
    \midrule
    \textbf{Mean} 
        & 85.1 & 83.2 & 87.5 & 92.0  
        & 84.9 & 87.8 & 88.6 & 93.3  
        & 86.1 & 88.8 & 89.6 & 94.0  \\
    \bottomrule
    \end{tabular}%
    }
    \vspace{-2mm}
\end{table*}

\newpage

\begin{table*}[h]
    \centering
    \caption{\small P-AUROC (\%) results of MVTec-AD for 1-shot, 2-shot, and 4-shot.}
    \vspace{-2mm}
    \label{table: Mvtec_P_AUROC_results}
    \resizebox{\textwidth}{!}{%
    \begin{tabular}{lcccccccccccc}
    \toprule
    \multirow{2}{*}{\textbf{Class}} 
    & \multicolumn{4}{c}{\textbf{1-shot}} 
    & \multicolumn{4}{c}{\textbf{2-shot}} 
    & \multicolumn{4}{c}{\textbf{4-shot}} 
    \\
    \cmidrule(lr){2-5}\cmidrule(lr){6-9}\cmidrule(lr){10-13}
        & \textbf{WinCLIP} & \textbf{PromptAD} & \textbf{IIPAD} & \textbf{Ours}  
        & \textbf{WinCLIP} & \textbf{PromptAD} & \textbf{IIPAD} & \textbf{Ours} 
        & \textbf{WinCLIP} & \textbf{PromptAD} & \textbf{IIPAD} & \textbf{Ours} \\
    \midrule
    Bottle      
        & 95.1  & 94.5  & 98.4  & 98.6  
        & 95.5  & 95.9  & 98.6  & 98.6 
        & 95.2  & 96.9  & 98.6  & 98.6 \\
    Cable       
        & 74.6  & 76.8  & 94.9  & 94.7 
        & 76.4  & 80.7  & 95.3  & 95.0
        & 77.1  & 83.7  & 96.4  & 94.7 \\
    Capsule     
        & 95.6  & 94.6  & 97.2  & 98.6 
        & 94.7  & 94.8  & 97.5  & 98.8 
        & 96.3  & 97.7  & 97.2  & 98.8 \\
    Carpet      
        & 99.1  & 99.1  & 99.5  & 99.5 
        & 99.1  & 99.2  & 99.5  & 99.4 
        & 99.1  & 99.2  & 99.5  & 99.5 \\
    Grid        
        & 95.1  & 96.8  & 96.6  & 99.2
        & 96.3  & 96.0  & 96.3  & 99.3 
        & 96.0  & 96.8  & 98.1  & 99.3  \\
    Hazelnut    
        & 98.6  & 96.9  & 98.4  & 99.5 
        & 98.7  & 98.0  & 98.7  & 99.5  
        & 98.7  & 98.1  & 98.8  & 99.5  \\
    Leather     
        & 99.3  & 99.3  & 99.4  & 99.4 
        & 99.3  & 99.4  & 99.3  & 99.4  
        & 99.4  & 99.4  & 99.5  & 99.4  \\
    Metal\_nut   
        & 77.9  & 94.2  & 94.4  & 92.5 
        & 76.1  & 94.8  & 95.4  & 94.3 
        & 79.5  & 93.2  & 94.8  & 96.3  \\
    Pill        
        & 93.9  & 92.3  & 96.6  & 94.5 
        & 94.1  & 94.3  & 97.0  & 94.6 
        & 94.4  & 95.3  & 96.9  & 95.0  \\
    Screw       
        & 96.7  & 95.7  & 96.0  & 98.4 
        & 97.0  & 96.2  & 96.5  & 98.7 
        & 96.5  & 97.2  & 96.9  & 98.7  \\
    Tile        
        & 92.2  & 94.3  & 96.9  & 97.2
        & 92.6  & 95.5  & 97.1  & 97.3 
        & 92.2  & 95.9  & 97.3  & 97.4  \\
    Toothbrush  
        & 95.1  & 99.0  & 97.4  & 99.2  
        & 84.5  & 99.2  & 97.4  & 99.2  
        & 96.8  & 99.2  & 97.9  & 99.4  \\
    Transistor  
        & 85.1  & 75.5  & 86.6  & 85.0 
        & 95.3  & 84.4  & 88.4  & 86.6 
        & 84.4  & 87.5  & 89.6  & 87.4  \\
    Wood        
        & 94.7  & 95.8  & 97.0  & 96.2 
        & 94.7  & 95.3  & 97.3  & 96.2 
        & 94.6  & 95.6  & 97.4  & 96.2  \\
    Zipper      
        & 92.9  & 97.6  & 96.0  & 99.0
        & 92.4  & 97.3  & 96.7  & 99.0 
        & 92.9  & 96.9  & 96.7  & 99.1  \\
    \midrule
    \textbf{Mean} 
        & 92.4  & 91.8  & 96.4  & 96.8
        & 92.4  & 92.2  & 96.7  & 97.0
        & 92.9  & 92.9  & 97.0  & 97.2 \\
    \bottomrule
    \end{tabular}%
    }
    \vspace{-2mm}
\end{table*}

\begin{table*}[h]
    \centering
    \caption{P-AUROC (\%) results of VisA for 1-shot, 2-shot, and 4-shot.}
    \vspace{-2mm}
    \label{table: VisA_P_AUROC_results}
    \resizebox{\textwidth}{!}{%
    \begin{tabular}{lcccccccccccc}
    \toprule
    \multirow{2}{*}{\textbf{Class}} 
    & \multicolumn{4}{c}{\textbf{1-shot}} 
    & \multicolumn{4}{c}{\textbf{2-shot}} 
    & \multicolumn{4}{c}{\textbf{4-shot}} 
    \\
    \cmidrule(lr){2-5}\cmidrule(lr){6-9}\cmidrule(lr){10-13}
       & \textbf{WinCLIP} & \textbf{PromptAD} & \textbf{IIPAD } & \textbf{Ours}  
       & \textbf{WinCLIP} & \textbf{PromptAD} & \textbf{IIPAD } & \textbf{Ours}  
       & \textbf{WinCLIP} & \textbf{PromptAD} & \textbf{IIPAD } & \textbf{Ours}  \\
    \midrule
    Candle      
        & 93.8 & 97.1 & 98.5 & 99.5  
        & 94.7 & 97.7 & 98.5 & 99.6  
        & 95.0 & 97.7 & 98.6 & 99.5  \\
    Capsules    
        & 93.2 & 96.7 & 97.1 & 99.8  
        & 93.0 & 97.2 & 97.3 & 99.8  
        & 93.2 & 97.4 & 97.9 & 99.8  \\
    Cashew      
        & 94.6 & 97.9 & 97.6 & 99.6  
        & 95.3 & 97.8 & 97.1 & 99.6  
        & 94.7 & 97.9 & 97.5 & 99.6  \\
    Chewinggum  
        & 98.9 & 99.2 & 99.6 & 99.9  
        & 98.9 & 99.0 & 99.5 & 99.9  
        & 98.9 & 99.1 & 99.5 & 99.8  \\
    Fryum       
        & 95.1 & 95.6 & 96.2 & 99.8  
        & 95.6 & 95.8 & 96.1 & 99.8  
        & 95.4 & 96.0 & 95.8 & 99.7  \\
    Macaroni1   
        & 95.6 & 97.6 & 97.5 & 99.5  
        & 96.7 & 98.8 & 97.6 & 99.6  
        & 97.0 & 98.6 & 97.8 & 99.6  \\
    Macaroni2   
        & 94.0 & 95.6 & 96.6 & 99.9  
        & 94.4 & 96.1 & 96.9 & 99.9  
        & 93.8 & 98.1 & 96.9 & 99.9  \\
    Pcb1        
        & 94.1 & 96.9 & 98.5 & 99.5  
        & 94.6 & 98.1 & 98.5 & 99.5  
        & 94.7 & 98.8 & 98.7 & 99.6  \\
    Pcb2        
        & 92.4 & 94.0 & 94.6 & 99.6  
        & 93.1 & 95.1 & 96.8 & 99.6  
        & 93.3 & 95.6 & 96.6 & 99.7  \\
    Pcb3        
        & 91.6 & 95.3 & 93.0 & 99.8  
        & 92.4 & 95.5 & 93.7 & 99.9  
        & 93.2 & 96.4 & 94.0 & 99.9  \\
    Pcb4        
        & 94.2 & 95.7 & 95.2 & 99.7  
        & 94.9 & 96.8 & 96.6 & 99.7  
        & 95.6 & 96.9 & 97.2 & 99.7  \\
    Pipefryum   
        & 97.9 & 98.6 & 98.3 & 99.7  
        & 97.8 & 98.8 & 98.5 & 99.7  
        & 97.8 & 98.9 & 98.5 & 99.7  \\
    \midrule
    \textbf{Mean} 
        & 94.6 & 96.3 & 96.9 & 99.7  
        & 95.1 & 96.9 & 97.2 & 99.7  
        & 95.2 & 97.2 & 97.4 & 99.7  \\
    \bottomrule
    \end{tabular}%
    }
    \vspace{-2mm}
\end{table*}

\begin{table*}[h]
    \centering
    \caption{\small PRO (\%) results of MVTec-AD for 1-shot, 2-shot, and 4-shot.}
    \vspace{-2mm}
    \label{table: Mvtec_PRO_results}
    \resizebox{\textwidth}{!}{%
    \begin{tabular}{lcccccccccccc}
    \toprule
    \multirow{2}{*}{\textbf{Class}} 
    & \multicolumn{4}{c}{\textbf{1-shot}} 
    & \multicolumn{4}{c}{\textbf{2-shot}} 
    & \multicolumn{4}{c}{\textbf{4-shot}} 
    \\
    \cmidrule(lr){2-5}\cmidrule(lr){6-9}\cmidrule(lr){10-13}
        & \textbf{WinCLIP} & \textbf{PromptAD} & \textbf{IIPAD} & \textbf{Ours}  
        & \textbf{WinCLIP} & \textbf{PromptAD} & \textbf{IIPAD} & \textbf{Ours} 
        & \textbf{WinCLIP} & \textbf{PromptAD} & \textbf{IIPAD} & \textbf{Ours} \\
    \midrule
    Bottle      
        & 84.5 & 89.1 & 94.9 & 96.2 
        & 85.8 & 91.6 & 95.2 & 95.8
        & 84.8 & 93.4 & 95.5 & 95.9 \\
    Cable       
        & 55.9 & 70.4 & 86.7 & 89.2 
        & 62.1 & 71.2 & 89.5 & 89.9
        & 61.5 & 75.8 & 91.2 & 89.6 \\
    Capsule     
        & 89.0 & 81.2 & 91.3 & 94.3
        & 84.2 & 82.0 & 92.6 & 94.9
        & 88.9 & 90.7 & 88.9 & 94.8 \\
    Carpet      
        & 96.4 & 96.8 & 98.1  & 98.3
        & 96.1 & 96.9 & 98.0  & 98.3 
        & 96.0 & 96.6 & 97.7  & 98.2 \\
    Grid        
        & 85.3 & 91.9 & 88.4  & 96.1
        & 88.1 & 89.2 & 87.7  & 96.6
        & 86.8 & 91.9 & 91.3  & 96.8 \\
    Hazelnut    
        & 92.5 & 90.9 & 93.8  & 93.0
        & 93.1 & 93.6 & 94.6  & 93.2 
        & 92.4 & 93.3 & 95.8  & 93.4 \\
    Leather     
        & 98.2 & 98.3 & 98.7  & 98.7
        & 98.3 & 98.8 & 98.5  & 98.7
        & 98.2 & 97.8 & 98.6  & 98.4 \\
    Metal\_nut   
        & 77.0 & 90.1 & 91.8  & 91.8
        & 75.6 & 90.7 & 92.3  & 93.3 
        & 79.5 & 89.8 & 93.4  & 95.3 \\
    Pill        
        & 88.9 & 90.1 & 94.8  & 97.0
        & 89.6 & 93.3 & 94.1  & 97.1 
        & 89.9 & 94.6 & 95.6  & 96.7 \\
    Screw       
        & 87.0 & 83.9 & 85.3  & 92.6 
        & 88.1 & 84.5 & 85.8  & 92.8
        & 86.6 & 89.1 & 87.5  & 93.3 \\
    Tile        
        & 78.7 & 89.8 & 90.4  & 95.4
        & 79.7 & 90.7 & 90.7  & 95.1
        & 78.2 & 91.4 & 91.2  & 94.9 \\
    Toothbrush  
        & 86.5 & 93.4 & 82.4  & 94.1
        & 85.6 & 93.1 & 82.9  & 95.2
        & 88.5 & 92.3 & 84.4  & 95.6 \\
    Transistor  
        & 63.5 & 59.4 & 67.4  & 63.3
        & 62.8 & 66.0 & 68.6  & 63.8
        & 62.6 & 69.4 & 71.2  & 65.7 \\
    Wood        
        & 86.9 & 92.8 & 94.2  & 96.0
        & 87.8 & 93.4 & 94.6  & 95.9
        & 88.3 & 93.1 & 94.7  & 96.1 \\
    Zipper      
        & 81.9 & 92.5 & 88.3  & 96.8 
        & 81.8 & 91.7 & 89.6  & 97.1
        & 83.0 & 91.2 & 90.8  & 97.0 \\
    \midrule
    \textbf{Mean} 
        & 83.5 & 83.6 & 89.8  & 92.8
        & 83.9 & 84.3 & 90.3  & 93.3
        & 84.4 & 84.7 & 91.2  & 93.5\\
    \bottomrule
    \end{tabular}%
    }
    \vspace{-2mm}
\end{table*}

\begin{table*}[h]
    \centering
    \caption{\small PRO (\%) results of VisA for 1-shot, 2-shot, and 4-shot.}
    \vspace{-2mm}
    \label{table: VisA_PRO_results}
    \resizebox{\textwidth}{!}{%
    \begin{tabular}{lcccccccccccc}
    \toprule
    \multirow{2}{*}{\textbf{Class}} 
    & \multicolumn{4}{c}{\textbf{1-shot}} 
    & \multicolumn{4}{c}{\textbf{2-shot}} 
    & \multicolumn{4}{c}{\textbf{4-shot}} 
    \\
    \cmidrule(lr){2-5}\cmidrule(lr){6-9}\cmidrule(lr){10-13}
       & \textbf{WinCLIP} & \textbf{PromptAD} & \textbf{IIPAD} & \textbf{Ours}  
       & \textbf{WinCLIP} & \textbf{PromptAD} & \textbf{IIPAD} & \textbf{Ours}  
       & \textbf{WinCLIP} & \textbf{PromptAD} & \textbf{IIPAD} & \textbf{Ours}  \\
    \midrule
    Candle      
        & 89.6 & 92.3 & 95.0 & 98.4  
        & 90.2 & 92.3 & 95.3 & 98.7  
        & 90.5 & 92.6 & 95.4 & 98.4  \\
    Capsules    
        & 62.1 & 82.7 & 83.6 & 99.0  
        & 61.8 & 82.1 & 83.9 & 99.0  
        & 61.9 & 77.0 & 85.2 & 99.0  \\
    Cashew      
        & 87.6 & 89.9 & 92.9 & 98.8  
        & 86.7 & 88.1 & 93.0 & 98.5  
        & 86.7 & 88.3 & 92.8 & 98.5  \\
    Chewinggum  
        & 82.7 & 84.9 & 92.5 & 99.0  
        & 83.0 & 84.1 & 93.4 & 99.1  
        & 82.7 & 83.2 & 93.7 & 98.9  \\
    Fryum       
        & 87.5 & 81.9 & 87.7 & 98.0  
        & 87.8 & 80.8 & 88.1 & 98.2  
        & 88.7 & 81.9 & 89.2 & 98.1  \\
    Macaroni1   
        & 85.6 & 88.6 & 89.3 & 98.0  
        & 89.8 & 90.8 & 90.1 & 98.3  
        & 90.1 & 93.5 & 89.9 & 98.3  \\
    Macaroni2   
        & 81.0 & 83.7 & 86.6 & 99.8  
        & 81.0 & 85.2 & 86.3 & 99.8  
        & 79.8 & 91.2 & 87.1 & 99.9  \\
    Pcb1        
        & 68.8 & 87.9 & 84.3 & 98.1  
        & 70.2 & 86.5 & 85.1 & 98.2  
        & 70.5 & 87.1 & 85.6 & 98.5  \\
    Pcb2        
        & 73.6 & 73.4 & 77.3 & 97.6  
        & 74.0 & 76.8 & 77.8 & 97.6  
        & 74.1 & 77.9 & 77.4 & 98.0  \\
    Pcb3        
        & 76.7 & 79.0 & 76.8 & 99.1  
        & 79.4 & 79.5 & 78.7 & 99.4  
        & 80.3 & 83.6 & 78.1 & 99.5  \\
    Pcb4        
        & 79.9 & 76.7 & 83.7 & 93.2  
        & 82.4 & 83.7 & 84.3 & 94.7  
        & 83.8 & 82.0 & 87.3 & 95.2  \\
    Pipe\_fryum   
        & 95.7 & 96.2 & 97.2 & 96.9  
        & 95.7 & 96.9 & 97.1 & 97.0  
        & 95.8 & 96.7 & 97.1 & 96.8  \\
    \midrule
    \textbf{Mean} 
        & 80.9 & 82.2 & 87.3 & 98.0  
        & 81.8 & 85.2 & 87.9 & 98.2  
        & 82.1 & 84.7 & 88.3 & 98.4  \\
    \bottomrule
    \end{tabular}%
    }
    \vspace{-2mm}
\end{table*}

\FloatBarrier

\newpage

\subsection{Class-Wise Qualitative Results}
\label{more_qualitative_results}
\begin{figure}[h]
  \centering
  \vspace{14px}
  \begin{overpic}[width=0.78\linewidth,trim=0cm 0 0 0cm,clip]{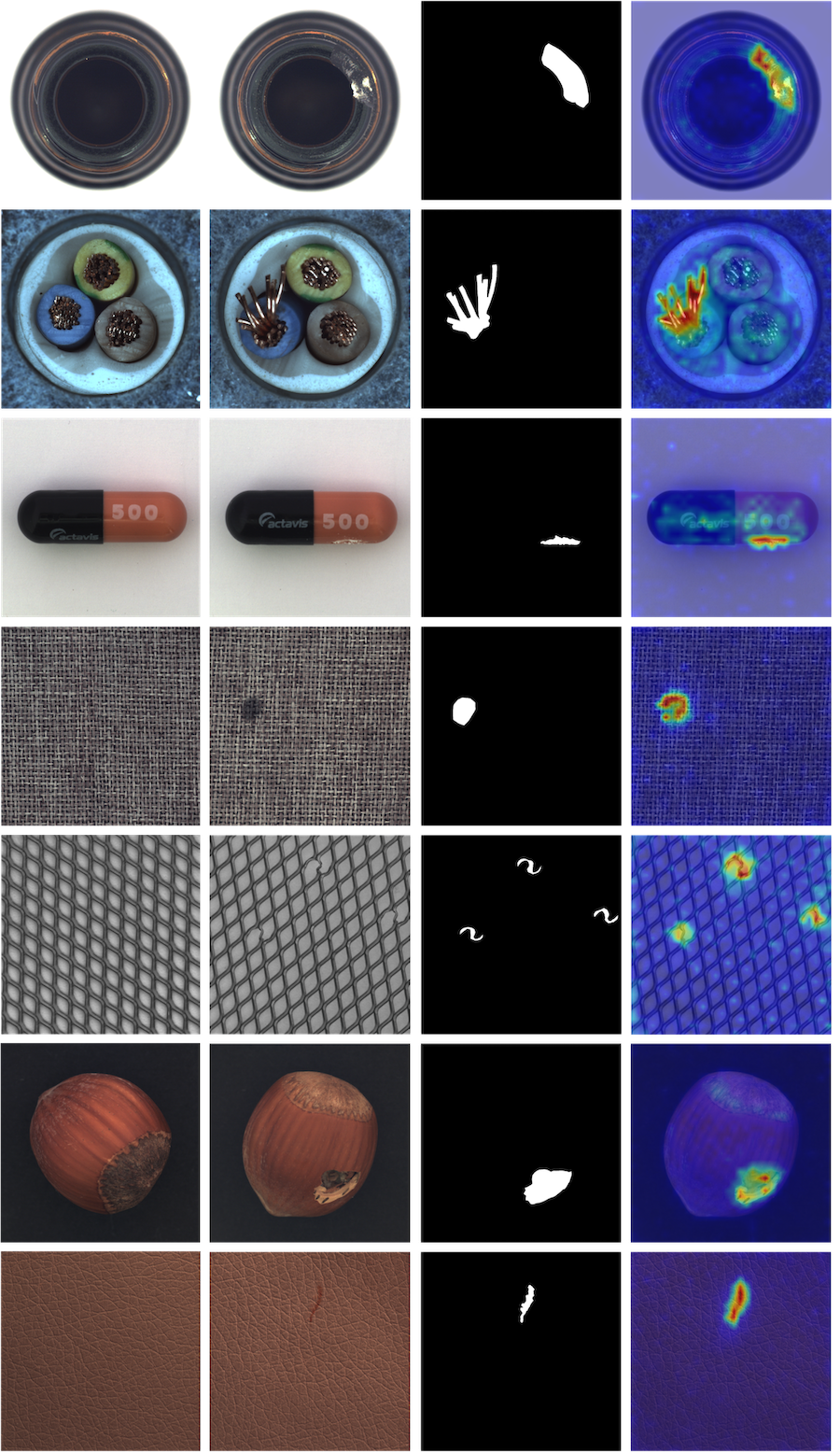}
    
    \put(1,101.5){\color{black}\small {Normal (Training)}}
    \put(15,101.5){\color{black}\small {Abnormal (Testing)}}
    \put(31.5,101.5){\color{black}\small {Ground Truth}}
    \put(46.6,101.5){\color{black}\small {\name{}}}
    
    \put(-2.5,92){\rotatebox{90}{\small Bottle}}
    \put(-2.5,76.5){\rotatebox{90}{\small Cable}}
    \put(-2.5,62){\rotatebox{90}{\small Capsule}}
    \put(-2.5,47.5){\rotatebox{90}{\small Carpet}}
    \put(-2.5,34){\rotatebox{90}{\small Grid}}
    \put(-2.5,18){\rotatebox{90}{\small Hazelnut}}
    \put(-2.5,4.5){\rotatebox{90}{\small Leather}}

  \end{overpic}
  \vspace{0.5px}
    \caption{\small More Qualitative 1-shot Results on MVTec-AD (i).
    }
    \label{fig:qualitative_comparison_mvtec_1}
    \vspace{-3mm}
\end{figure}

\begin{figure}[t]
  \centering
  \begin{overpic}[width=0.78\linewidth,trim=0cm 0 0 0cm,clip]{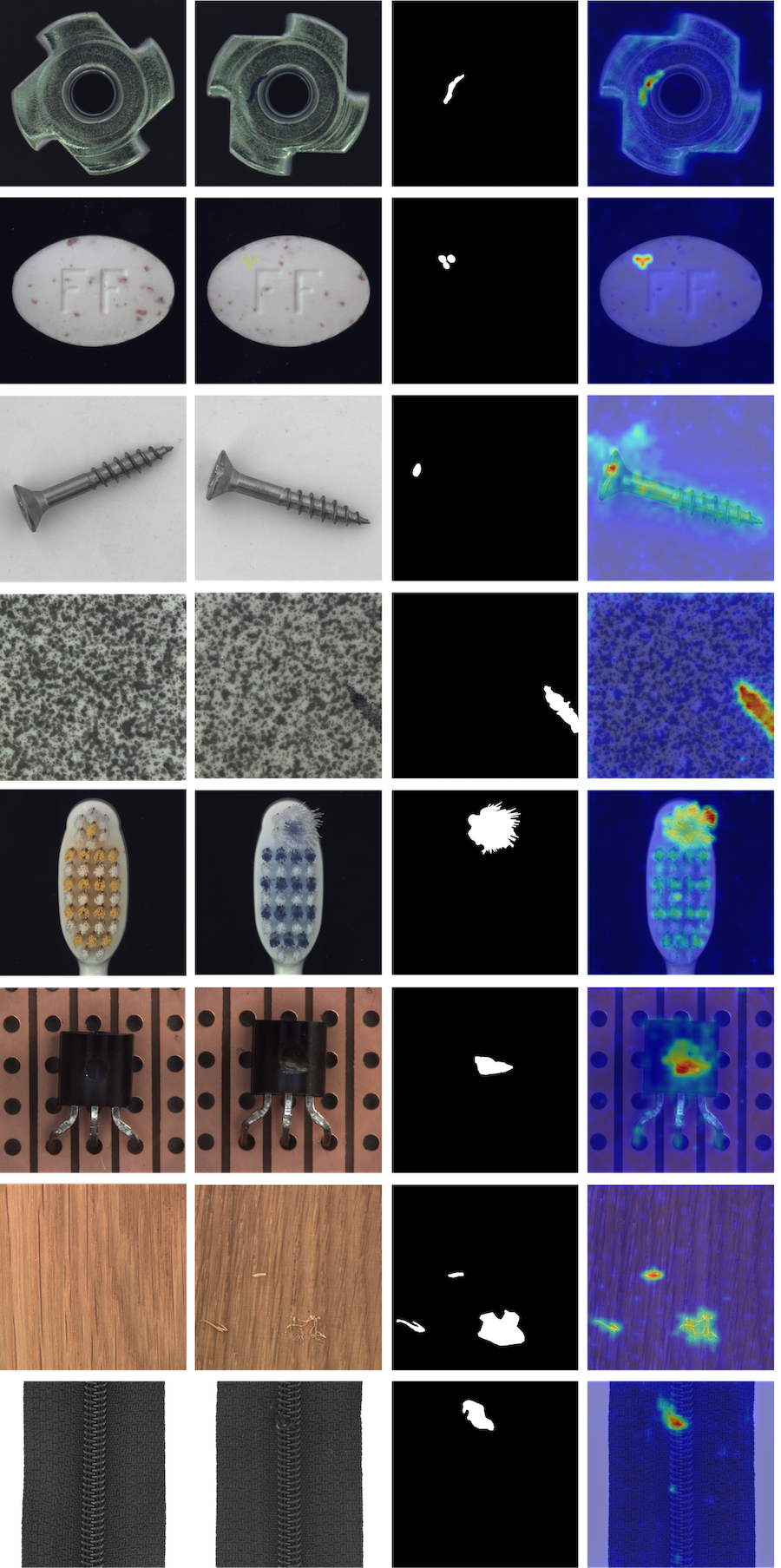}
    
    \put(1,101.5){\color{black}\small {Normal (Training)}}
    \put(12.7,101.5){\color{black}\small {Abnormal (Testing)}}
    \put(27,101.5){\color{black}\small {Ground Truth}}
    \put(40,101.5){\color{black}\small {\name{}}}
    
    \put(-2.5,91){\rotatebox{90}{\small Metal Nut}}
    \put(-2.5,80.5){\rotatebox{90}{\small Pill}}
    \put(-2.5,67){\rotatebox{90}{\small Screw}}
    \put(-2.5,55){\rotatebox{90}{\small Tile}}
    \put(-2.5,40.5){\rotatebox{90}{\small Toothbrush}}
    \put(-2.5,28){\rotatebox{90}{\small Transistor}}
    \put(-2.5,17){\rotatebox{90}{\small Wood}}
    \put(-2.5,4.5){\rotatebox{90}{\small Zipper}}

  \end{overpic}
  \vspace{0.5px}
    \caption{\small More Qualitative 1-shot Results on MVTec-AD (ii).
    }
    \label{fig:qualitative_comparison_mvtec_2}
    \vspace{-3mm}
\end{figure}

\begin{figure}[t]
  \centering
  \begin{overpic}[width=0.8\linewidth,trim=0cm 0 0 0cm,clip]{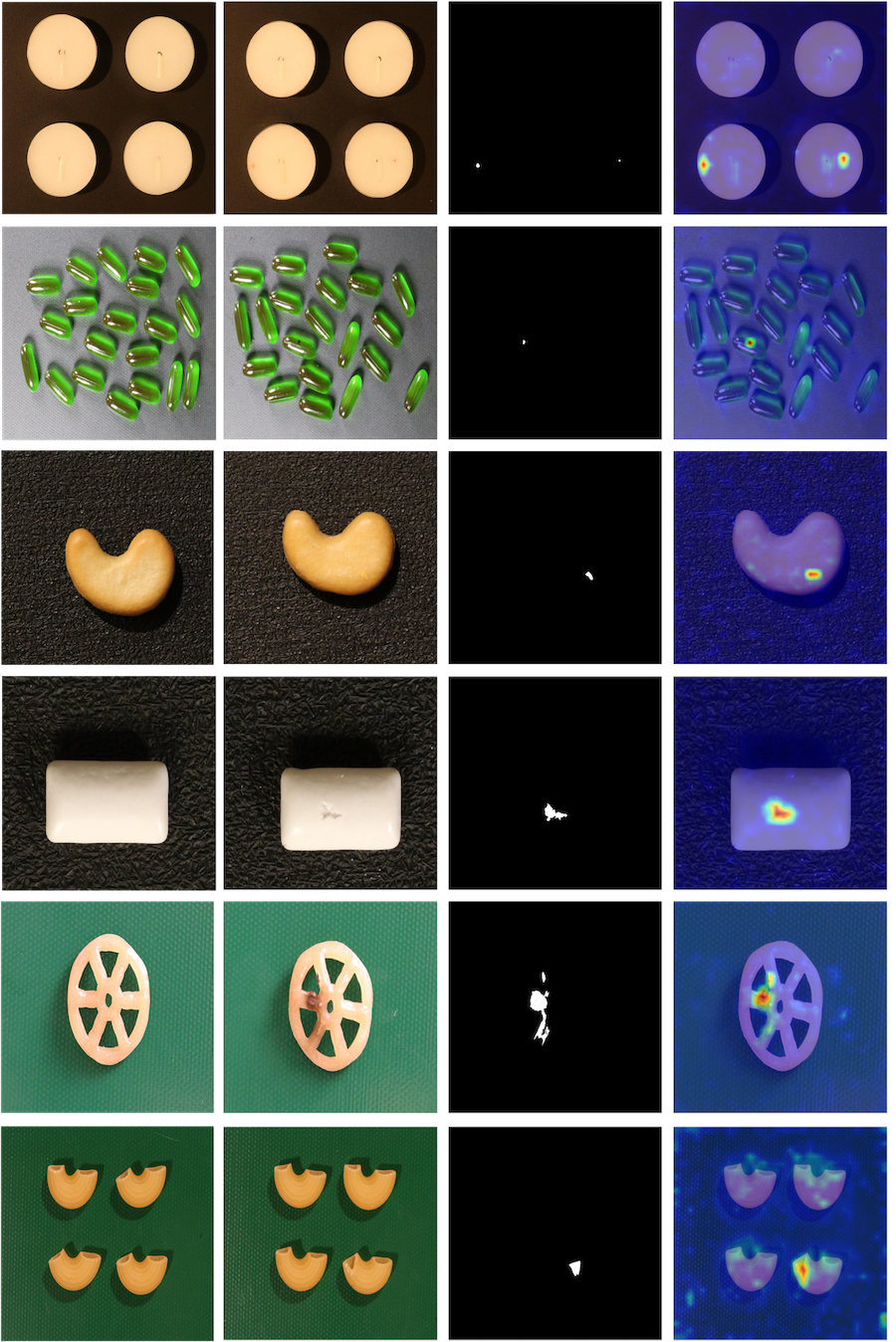}
    
    \put(1,101.5){\color{black}\small {Normal (Training)}}
    \put(17,101.5){\color{black}\small {Abnormal (Testing)}}
    \put(36,101.5){\color{black}\small {Ground Truth}}
    \put(53.5,101.5){\color{black}\small {\name{}}}
    
    \put(-2.5,90){\rotatebox{90}{\small Candle}}
    \put(-2.5,71.5){\rotatebox{90}{\small Capsules}}
    \put(-2.5,56){\rotatebox{90}{\small Cashew}}
    \put(-2.5,37){\rotatebox{90}{\small Chewinggum}}
    \put(-2.5,23){\rotatebox{90}{\small Fryum}}
    \put(-2.5,4){\rotatebox{90}{\small Macaroni1}}

  \end{overpic}
  \vspace{0.5px}
    \caption{\small More Qualitative 1-shot Results on VisA (i).
    }
    \label{fig:qualitative_comparison_Visa_1}
    \vspace{-3mm}
\end{figure}

\begin{figure}[t]
  \centering
  \begin{overpic}[width=0.8\linewidth,trim=0cm 0 0 0cm,clip]{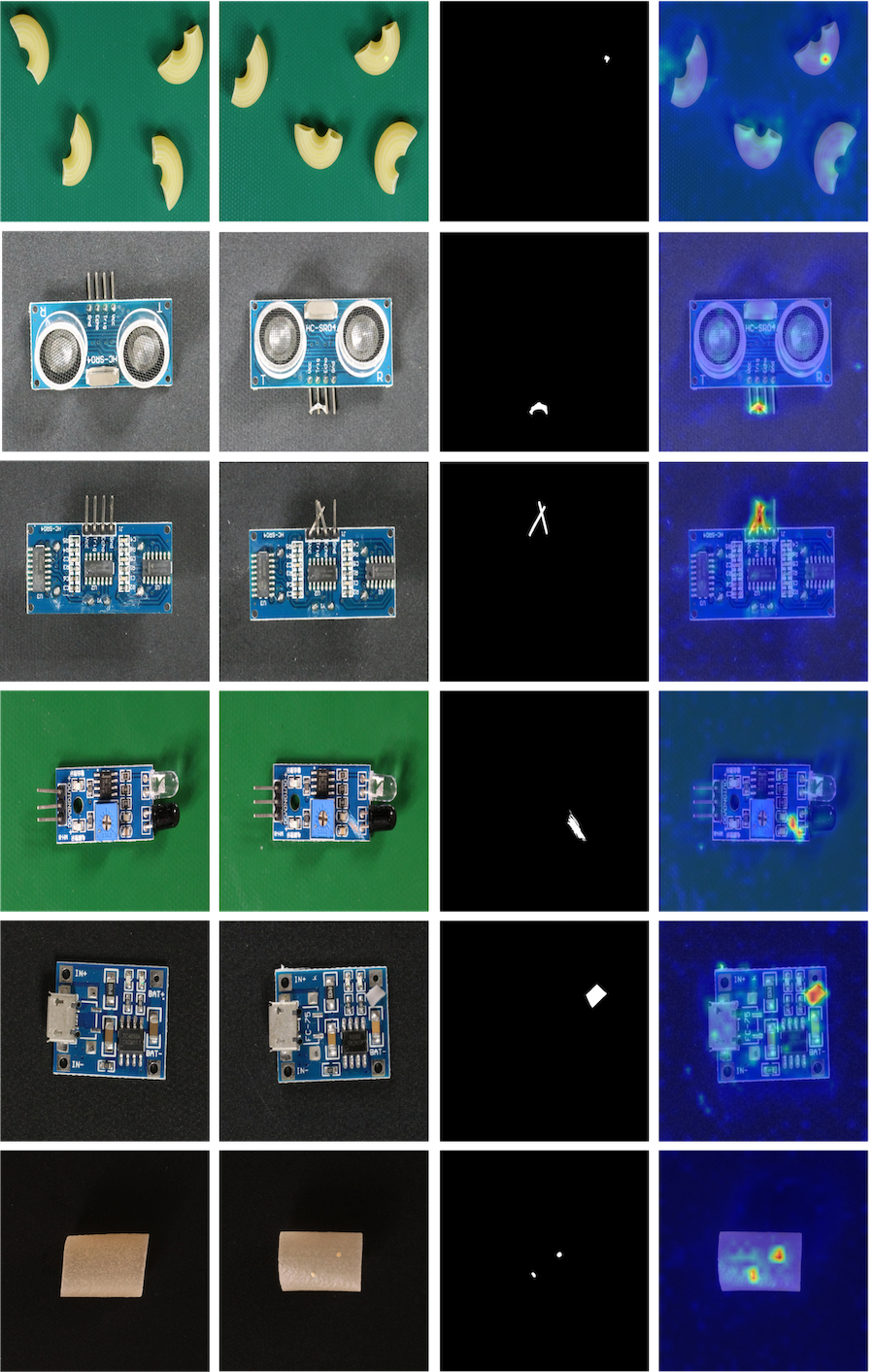}
    
    \put(1,101.5){\color{black}\small {Normal (Training)}}
    \put(17,101.5){\color{black}\small {Abnormal (Testing)}}
    \put(34.5,101.5){\color{black}\small {Ground Truth}}
    \put(51,101.5){\color{black}\small {\name{}}}
    
    \put(-2.5,88){\rotatebox{90}{\small Macaroni2}}
    \put(-2.5,73){\rotatebox{90}{\small Pcb1}}
    \put(-2.5,56.8){\rotatebox{90}{\small Pcb2}}
    \put(-2.5,39.9){\rotatebox{90}{\small Pcb3}}
    \put(-2.5,23){\rotatebox{90}{\small Pcb4}}
    \put(-2.5,4){\rotatebox{90}{\small Pipe Fryum}}

  \end{overpic}
  \vspace{0.5px}
    \caption{\small More Qualitative 1-shot Results on VisA (ii).
    }
    \label{fig:qualitative_comparison_Visa_2}
    \vspace{-3mm}
\end{figure}

The class-wise qualitative results of our method in 1-shot setting on MVTec-AD and VisA are shown in \autoref{fig:qualitative_comparison_mvtec_1}, \autoref{fig:qualitative_comparison_mvtec_2}, \autoref{fig:qualitative_comparison_Visa_1}, and \autoref{fig:qualitative_comparison_Visa_2}. We show each training sample in the first column.

\newpage

\newpage

\newpage

\end{document}